\documentclass{article}

\usepackage[preprint]{neurips_2025}

\usepackage[utf8]{inputenc} 
\usepackage[T1]{fontenc}    
\usepackage{hyperref}       
\usepackage{url}            
\usepackage{booktabs}       
\usepackage{amsfonts}       
\usepackage{nicefrac}       
\usepackage{microtype}      
\usepackage{xcolor}         

\usepackage{graphicx}
\usepackage{algorithm}
\usepackage{algorithmic}
\usepackage{natbib}
\usepackage{subfigure}
\usepackage{bbm}
\usepackage{listings}
\definecolor{darkgreen}{rgb}{0,0.5,0}
\usepackage{amsmath}
\usepackage{wrapfig}

\usepackage{color, soul}
\usepackage[most,breakable]{tcolorbox}
\usepackage{xcolor}         

\usepackage[labelfont=bf]{caption}
\usepackage{enumitem}
\usepackage{sectsty}

\newtcolorbox{mytheorem}{
  colback=gray!5, 
  colframe=gray!80, 
  boxrule=0.5pt, 
  arc=4pt, 
  left=4pt, 
  right=4pt, 
  top=4pt, 
  bottom=4pt, 
}
\title{The Climb Carves Wisdom Deeper Than the Summit:\\ On the Noisy Rewards in Learning to Reason}

\author{%
  Ang Lv$^{1}$
  \quad Ruobing Xie$^{2}$\thanks{Corresponding authors} 
  \quad Xingwu Sun$^{2,3}$
  \quad Zhanhui Kang$^{2}$ 
  \quad Rui Yan$^{1,4*}$ \\
  $^{1}$ GSAI, Renmin University of China \\
  $^{2}$ Large Language Model Department, Tencent \\
  $^{3}$ University of Macau \\
  $^{4}$ School of Computer Science, Wuhan University \\
  \texttt{\{anglv, ruiyan\}@ruc.edu.cn}  \\
\texttt{\{ruobingxie, sammsun, kegokang\}@tencent.com }\\
}

\begin{document}

\maketitle

\begin{abstract}
Recent studies on post-training large language models (LLMs) for reasoning through reinforcement learning (RL) typically focus on tasks that can be accurately verified and rewarded, such as solving math problems. 
In contrast, our research investigates the impact of reward noise, a more practical consideration for real-world scenarios involving the post-training of LLMs using reward models.
We found that LLMs demonstrate strong robustness to substantial reward noise. 
For example, manually flipping 40\% of the reward function's outputs in math tasks still allows a Qwen-2.5-7B model to achieve rapid convergence, improving its performance on math tasks from 5\% to 72\%, compared to the 75\% accuracy achieved by a model trained with noiseless rewards.
Surprisingly, by only rewarding the appearance of key reasoning phrases (namely reasoning pattern reward, RPR), such as ``first, I need to''—without verifying the correctness of answers, the model achieved peak downstream performance (over 70\% accuracy for Qwen-2.5-7B) comparable to models trained with strict correctness verification and accurate rewards. 
Recognizing the importance of the reasoning process over the final results, we combined RPR with noisy reward models. 
RPR helped calibrate the noisy reward models, mitigating potential false negatives and enhancing the LLM's performance on open-ended tasks. 
These findings suggest the importance of improving models' foundational abilities during the pre-training phase while providing insights for advancing post-training techniques. Our code and scripts are available at \url{https://github.com/trestad/Noisy-Rewards-in-Learning-to-Reason}.
\end{abstract}

\section{Introduction}

Reinforcement learning (RL) applied to post-training large language models (LLMs) has led to significant advancements in enhancing their thinking and reasoning abilities~\citep{deepseekai2025deepseekr1incentivizingreasoningcapability,kimi}, resulting in improved performance on many challenging downstream tasks. 
Most current research focuses on math tasks~\citep{orz,tinyzero,yeo2025demystifyinglongchainofthoughtreasoning,gandhi2025cognitivebehaviorsenableselfimproving,overthink}, as these can be easily verified as correct or incorrect by simple rule-based reward functions. 
However, in many real-world applications, such as preference alignment~\citep{ouyang,pmlr-v202-zhu23f} and open question-answering~\citep{jaques-etal-2020-human,nakano2022webgptbrowserassistedquestionansweringhuman}, responses cannot be easily quantified with simple rule-based functions and instead require evaluation by neural reward models. 
These models, being imperfect, often introduce noise even resulting in opposite rewards.
In this study, we studied scenarios in which the reward, whether derived from a neural reward model or a rule-based function, contains noise, aiming to gain a more practical understanding of noisy rewards in teaching LLMs to reason.

\begin{wrapfigure}{r}{0.45\linewidth}
    \includegraphics[width=\linewidth]{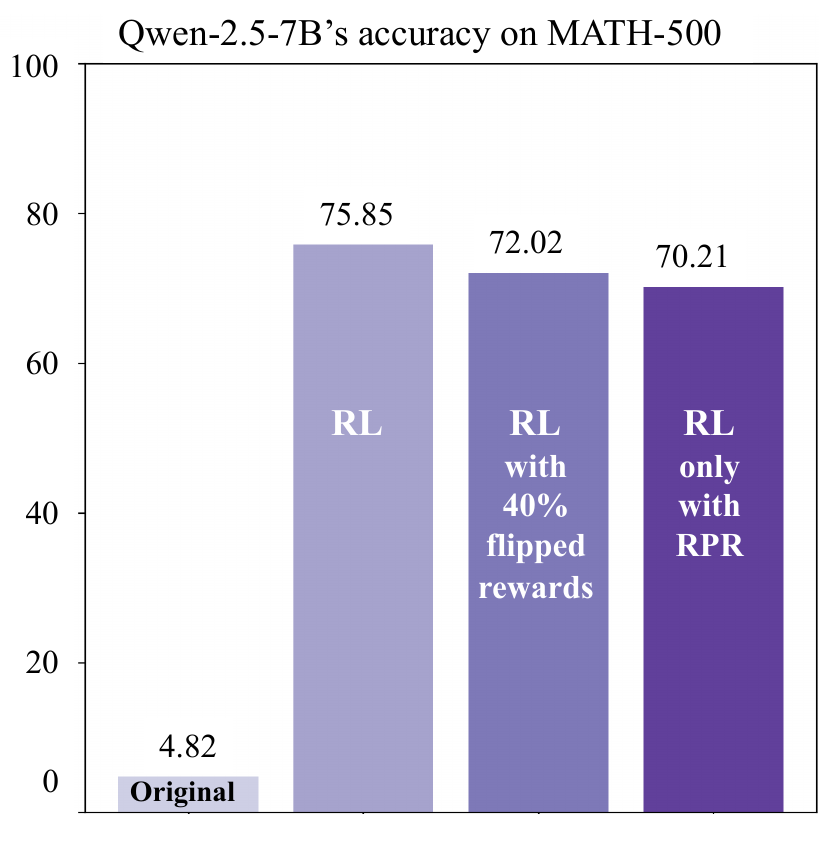}
    \caption{All of (1) standard RL, (2) RL with 40\% of the rewards manually flipped to the opposite, and (3) RL with only Reasoning Pattern Rewards (RPR) (i.e., rewards are given whenever key reasoning phrases appear, without verifying the final answer)—can improve Qwen-2.5-7B's accuracy on MATH-500 from an initial 5\% to over 70\%. 
    The performance gap between these three setups is minimal compared to the overall improvements. 
    }
    \label{fig:teaser}
\end{wrapfigure}
We made an unexpected discovery: despite the presence of substantial noise in the rewards, the model can still be effectively trained and achieve fast convergence. 
For example, when training on math problems and introducing noise by randomly flipping 40\% of the reward function's outputs (i.e., assigning positive rewards for incorrect answers), a Qwen-2.5-7B~\citep{qwen2.5} model still improved its MATH-500~\citep{math} score from an initial 5\% to a surprising peak of 72.02\%, close to the 75.85\% achieved using a noiseless reward function. 
Training collapses when the flip rate reaches 50\%, where the reward becomes entirely random. 
The model's ability to tolerate substantial noisy rewards suggests that, although outputs with incorrect answers are mistakenly rewarded, they still exhibit valuable logical reasoning processes. 
These reasoning patterns are also valuable.
If this were not the case, a high frequency of incorrect rewards would likely hinder performance or, at the very least, slow convergence.
Thus, we hypothesize that the effectiveness of RL in enhancing LLMs primarily stems from the exploration of appropriate reasoning patterns during rollouts. 
Through this exploration, the model can more effectively leverage its pretrained knowledge by adopting strong reasoning patterns, which increases the likelihood of arriving at the correct answer.\footnote{A premise of this hypothesis is that the model must have acquired substantial reasoning capabilities during pretraining. As we show later, Qwen significantly outperforms Llama~\citep{llama3} not only in downstream performance after RL but also in robustness to noisy rewards—a difference that aligns with Llama's widely recognized weakness in reasoning~\citep{tinyzero,yeo2025demystifyinglongchainofthoughtreasoning,gandhi2025cognitivebehaviorsenableselfimproving}.}

To support this hypothesis, we conducted another experiment in which the model was rewarded whenever key reasoning phrases, such as ``first, I need to'' or ``let me first,'' appeared in the outputs, without verifying the correctness of the final answer.
We name this strategy as Reasoning Pattern Reward (RPR).
Using only RPR, the model achieved peak task performance (70.21\% for Qwen-2.5-7B) comparable to the performance achieved when the correctness of the solution was strictly verified. 
This provides strong evidence suggesting that LLMs can reason through RL because they have already learned to reason during pretraining, as no correctness supervision signals were given, meaning no new knowledge was learned.

LLMs' robustness to noisy rewards extends to open-ended NLP tasks as well.
We conducted experiments using the NVIDIA HelpSteer3 dataset~\citep{helpsteer}, which consists of a broad range of challenging open-ended questions requiring AI assistance. 
We varied reward model accuracy by adjusting training set size and found that LLMs trained with a 75\% accurate reward model performed comparably to those using our best model (85\% accuracy).

These insights motivate our proposal of a simple yet effective method to improve LLM performance on open-ended NLP tasks post-trained with noisy reward models.
We use RPR to calibrate the reward models by compensating for potential false negative signals, resulting in up to a 30\% net win rate over LLMs post-trained with vanilla reward models. 
RPR-calibrated reward models also enable smaller models, such as Qwen-2.5-3B, to demonstrate strong reasoning capabilities on complex open-ended tasks, where vanilla reward models lead to training collapse.

In summary, we provide some insights into training LLMs to reason via RL with noisy rewards:

1. We demonstrate that LLMs with strong inherent reasoning abilities are surprisingly robust to reward noise.

2. We present direct evidence showing that after RL, models' enhanced performance on challenging tasks primarily stems from adapted output patterns, which are more likely to lead to correct answers, rather than from learning much new knowledge during RL.

3. We propose a simple yet effective method to calibrate noisy reward models by rewarding reasoning patterns, leading to improved LLM performance on open-ended NLP tasks and unlocking smaller models' reasoning capabilities.

\section{Noisy verification rewards in math tasks}
\label{sec:math-exp}
Mathematics is one of the most commonly studied domains in LLM reasoning, due to its straightforward rule-based reward and evaluation. 
To explore the RL performance under noisy rewards, we begin by manually introducing noise to RL rewards in math tasks.
While math tasks are typically assumed to be noiseless and verification is considered accurate, our noisy scenario is practical for two reasons: (1) false negative rewards are common in math tasks, where a correct answer may be flagged as incorrect due to formatting issues, and (2) many proof-based questions in math tasks are difficult to verify accurately.

\subsection{Settings}

\textbf{Training.} Most of the training setups follow the approach in~\citep{orz}, which provides a simplified framework designed to help LLMs learn to reason. 
The experiments are based on VeRL~\citep{sheng2024hybridflow} framework by Volcengine.

\begin{figure}[h]
    \centering
    \includegraphics[width=\linewidth]{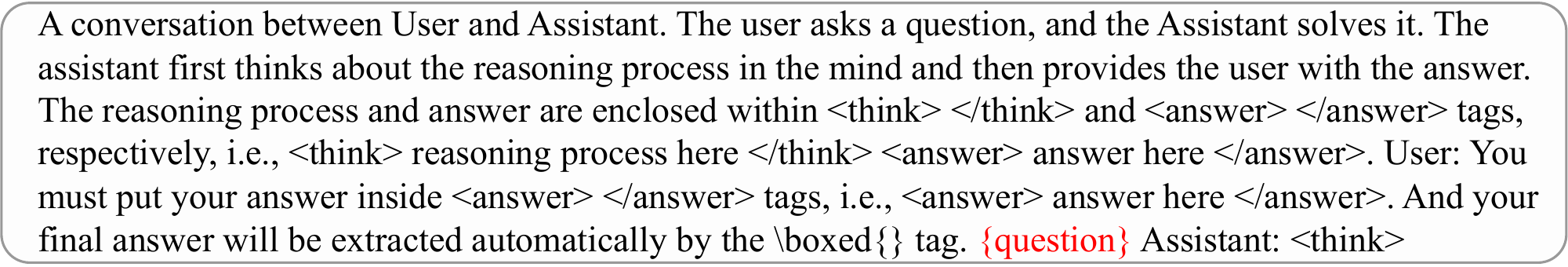}
    \caption{The prompt used in math training, where the ``question'' placeholder will be replaced with a specific question.}
    \label{fig:math-prompt}
\end{figure}

Specifically: the dataset includes 57K high-quality, source-mixture math problems spanning various difficulty levels. 
The prompts used in these tasks are shown in Figure~\ref{fig:math-prompt}. 
The outputs are extracted from the box tag, normalized for format (e.g., converting fractions to decimals and translating LaTeX answers to plain text), and compared to the ground truth. 
In the absence of manual reward noise, the model is given a reward of 1 if the output matches the ground truth and 0 otherwise.
For training, we employ vanilla PPO~\citep{ppo} with GAE~\citep{gae}, using $\lambda$ = 1 and $\gamma$ = 1, with no KL-regularization. 
The training batch size is 128, with a maximum response length of 4096.
The learning rate for the actor is $10^{-6}$, and for the critic, it is $5\times10^{-6}$.
To ensure stable training, we apply critic warmups for 20 steps, initially training the critic before training the actor model. 
We set the rollout number to 4. 
The primary focus of the main text is the Qwen-2.5-7B~\citep{qwen2.5}, which has demonstrated strong reasoning potential~\citep{gandhi2025cognitivebehaviorsenableselfimproving,yeo2025demystifyinglongchainofthoughtreasoning}, while experimental results on other models are provided in the appendix.

\textbf{Evaluation.} We use three datasets—MATH-500~\citep{math}, GPQA~\citep{gpqa}, and AIME 2024~\citep{aime}—to assess the model's reasoning ability on challenging tasks. 
We report the Pass@1 accuracy dynamics throughout the training.

\textbf{Random reward flip.}
We train the model by randomly flipping the reward with a probability $p$, where a reward of 1 is transformed to 0, and vice versa. 
This flip is applied on a question-wise basis, meaning that if a reward flip occurs for a given question, the rewards for all rollout outputs corresponding to that question will be flipped. 
Note that flipping rewards on an output-wise basis does not effectively introduce noise. 
For instance, when an LLM generates multiple correct outputs, some of which are rewarded correctly while others are not, it results in a sparse reward distribution. 
Such sparsity can only slow convergence and has minimal impact on the model's final performance.

\subsection{Experiments}

\begin{figure}
    \centering
    \includegraphics[width=\linewidth]{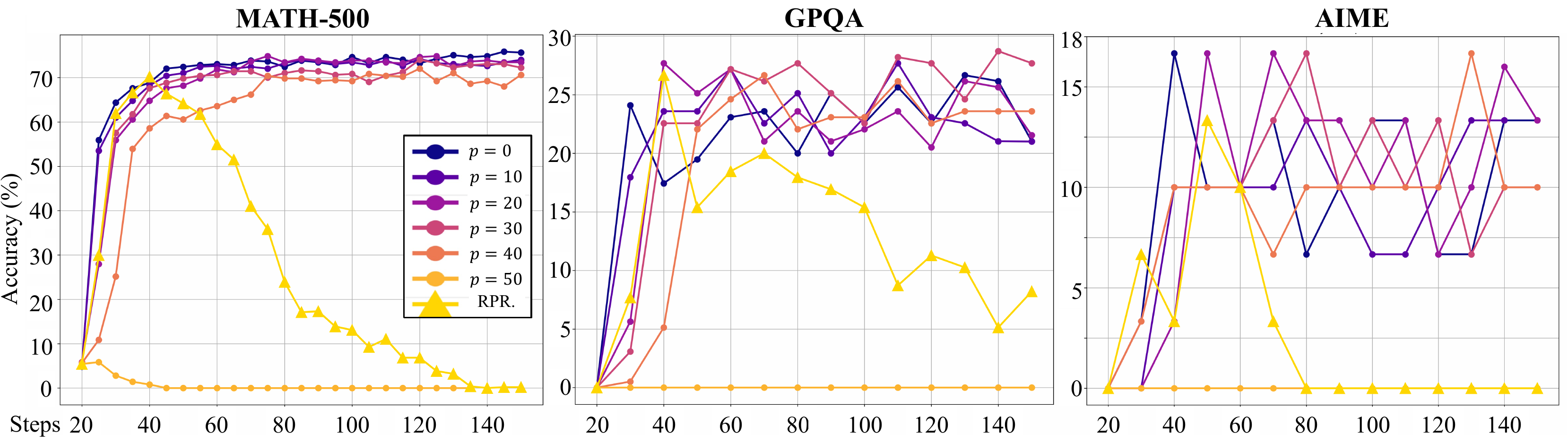}
    \caption{Accuracy on three test sets during training.
Due to critic warmup, the actor model is not updated during the first 20 steps; thus, the x-axis begins at step 20.}
    \label{fig:noise-p}
\end{figure}

\textit{\textbf{Experiment 1.}}
We train the model with the probability $p$ of noise increasing from 0\% to 50\%, with intervals of 10\%, corresponding to increasingly random reward flips. 
The results are shown in Figure~\ref{fig:noise-p}. 
We only display the first 150 steps, as the performance has already plateaued.

In MATH-500, it is evident that even with a high flip rate of 40\%, the final performance is only slightly lower than that of the model trained with no noise (peak score of 72.02\% versus 75.85\%). 
For lower noise levels, the final performance is comparable, and the convergence also occurs at a similar rate. 
Only when we increase $p$ to 50\%, leading the reward entirely random, the training collapse.

In the other two tasks, similar trends are observed, though fluctuations are more pronounced. In particular, for $p$-20\%, the peak performance is even higher than when $p$=0.
We attribute such fluctuations to the inherent difficulty of the datasets, which cause larger variability in model performance.
In Appendix~\ref{apx:other-models}, we also demonstrate that Qwen-2.5-3B exhibits strong robustness to noise, comparable to the 7B models. 
In contrast, Llama-3.1-8B shows considerably weaker reasoning capabilities and underperforms even in noiseless settings.
This underperformance aligns with prior findings that highlight the limited foundational reasoning abilities and RL improvements in Llama models~\citep{tinyzero,yeo2025demystifyinglongchainofthoughtreasoning,gandhi2025cognitivebehaviorsenableselfimproving}. 
These limitations result in poor noise tolerance.

Based on the overall performance of the three models analyzed in this study, we carefully formulate Takeaway 1 with an important caveat: while large language models can exhibit robustness to reward noise, this property holds primarily for models with strong reasoning capabilities. 
This is both a practical and necessary condition—models lacking foundational reasoning abilities, such as Llama-3.1-8B, are unlikely to be effectively used in real-world applications, regardless of the level of reward noise, and are therefore of limited interest for our study.
\begin{mytheorem}
\textit{\textbf{Takeaway 1.}} For models with strong reasoning potential, even with significant opposite noise in the \textit{verification} rewards, the model can still be effectively trained during RL.
\end{mytheorem}

\textit{\textbf{Experiment 2.}}
Given the surprising result from Experiment 1, the key question is why assigning a reward of 1 to outputs with genuinely incorrect answers does not have a significant detrimental effect. 
Since the answer is incorrect, we hypothesize that the reasoning process itself might still be valuable and worth rewarding.
If this were not the case, it would be challenging to achieve performance comparable to noiseless setups, and convergence would likely be slower at the very least.

To test this hypothesis, we conducted the experiment 2:
We first identified $n$ high-frequency phrases that imply certain desired reasoning patterns, such as \emph{``We know that''} and \emph{``First I need to,''} in the outputs of a model trained with $p$=0.\footnote{These phrases also frequently appear in model outputs trained with higher levels of noise.}
Next, we designed a rule-based reward function: instead of verifying the correctness of the answer, the model would receive a reward of value $r$ each time a pre-identified reasoning phrase appeared in the output. 
The total reward is clipped to 1, creating a simple keyword-matching reward that ranged from 0 to 1.
We name this strategy as \textbf{\textit{Reasoning Pattern Reward}} (RPR).

\begin{wrapfigure}{r}{0.48\linewidth}
    \includegraphics[width=\linewidth]{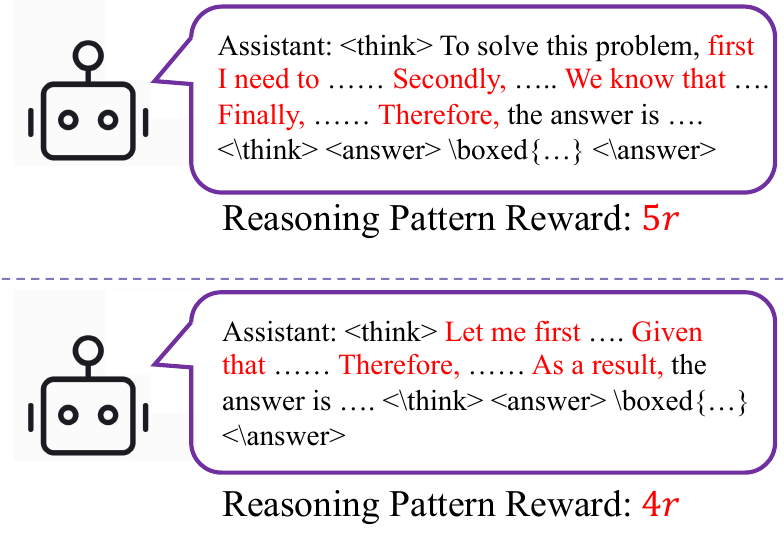}
    \caption{An illustration of how the reasoning pattern reward works through two example outputs.
Suppose the red text represents high-frequency phrases that we have pre-identified as indicating key reasoning processes.
In the first output, five key phrases are present, so the reward is 5$r$.
Similarly, the second output contains four key phrases, so the reward is 4$r$.
We do not verify the correctness of the answer.}
    \label{fig:example-think-reward}
\end{wrapfigure}
Figure~\ref{fig:example-think-reward} illustrates how RPR works, and an example code is provided in Figure~\ref{fig:code}.
To prevent the model from hacking the reward by outputting repeated reasoning phrases (e.g., ``We know that We know that We know that...''), a repetition penalty~\citep{yeo2025demystifyinglongchainofthoughtreasoning} is used. 
Since our goal was to validate the hypothesis rather than achieving stable state-of-the-art accuracies, $n$ was somewhat arbitrarily set to 40 without value tuning, and for simplicity, we set $r=1/n=0.025$.

The results are presented in Figure~\ref{fig:noise-p} for ease of comparison. 
Remarkably, even \emph{\textbf{without}} verifying the correctness of reasoning during the training process, the model demonstrates strong reasoning capabilities in the early stages, achieving a peak performance of 70.21\% on the MATH-500 task.
The peak performance on other tasks also shows a minimal gap compared to models trained without noise. 
However, as training progresses, the performance eventually declines. 
Upon analyzing the outputs, we found that this decline was due to overthinking.
Specifically, after reasoning through a few steps, the model revisits its prior thoughts, and this reasoning-then-revisiting cycle continues for too many iterations, resulting in an excessively long chain of reasoning that exceeds the context limit and is truncated before the final answer can be generated.
The model produces these long reasoning steps from multiple viewpoints, effectively ``escaping'' the repetition penalty. 
An example of this is shown in Figure~\ref{fig:overthink-case}, where the model has already arrived at the correct answer but continues to reason, preventing the extraction of the final answer.

The lack of answer verification and supervision, combined with strong peak performance, suggests that the model does not require substantial new knowledge through supervision to final answers in RL post-training. 
Instead, most of the improvements in solving challenging tasks were already learned during pretraining and are activated by RL through rewarding effective reasoning patterns that can lead to correct answers.

This experiment provides direct evidence illustrating the role of RL in LLMs during post-training.

\begin{mytheorem}
\textit{\textbf{Takeaway 2.}} Training LLMs solely based on reasoning pattern rewards, without any correctness check, can develop powerful, though transient, reasoning abilities.
This is a direct evidence that LLMs do not require much new knowledge during RL; instead, RL explores outputs that are likely to lead to correct answers and reinforces those reasoning patterns. 
\end{mytheorem}

\textbf{\textit{Remark.}} 
Key reasoning phrases we collected are broadly applicable across tasks. 
As shown in Experiment 4 (Section~\ref{sec:nlp-exp}), RPR patterns from math tasks remain effective in diverse, open-ended domains.
While there may be other effective keywords or even implicit hidden states~\citep{hao2024traininglargelanguagemodels} that trigger reasoning, pursuing a more refined design is not the focus of this study.

\section{Noisy reward models in open NLP tasks}
\label{sec:3}
Now we turn to the open NLP tasks requiring reward models (RMs).
Different from manually flipping rewards in math tasks, the noise level in open NLP tasks can be approximately reflected in the varying evaluation accuracies of RMs.
We first introduce the data, RM training details, and then introduce experiments with noisy RL rewards and corresponding findings.

\subsection{Preliminaries: Training reward models with varying accuracy}

\textbf{Dataset.}
We use the NVIDIA HelpSteer3~\citep{helpsteer} dataset, which contains 40.5K multi-domain open-ended questions that require helpful assistance. 
Each question is paired with two responses, evaluated by multiple annotators for helpfulness, categorized into seven fine-grained levels.
There is also a chat history preceding the current question, providing context for the question. 
The dataset is split into a training set of 38.5K samples and a validation set of 2K samples.

\textbf{Training.} Our reward model is built on a Qwen-2.5-7B model with an added prediction head. 
We simplify the original seven-level helpfulness scale into a binary classification task: the more helpful response in each pair is labeled as 1, and the less helpful one as 0.
For each response, we concatenate it with the chat history as the input to the reward model.
\begin{wrapfigure}{r}{0.45
\linewidth}
    \includegraphics[width=\linewidth]{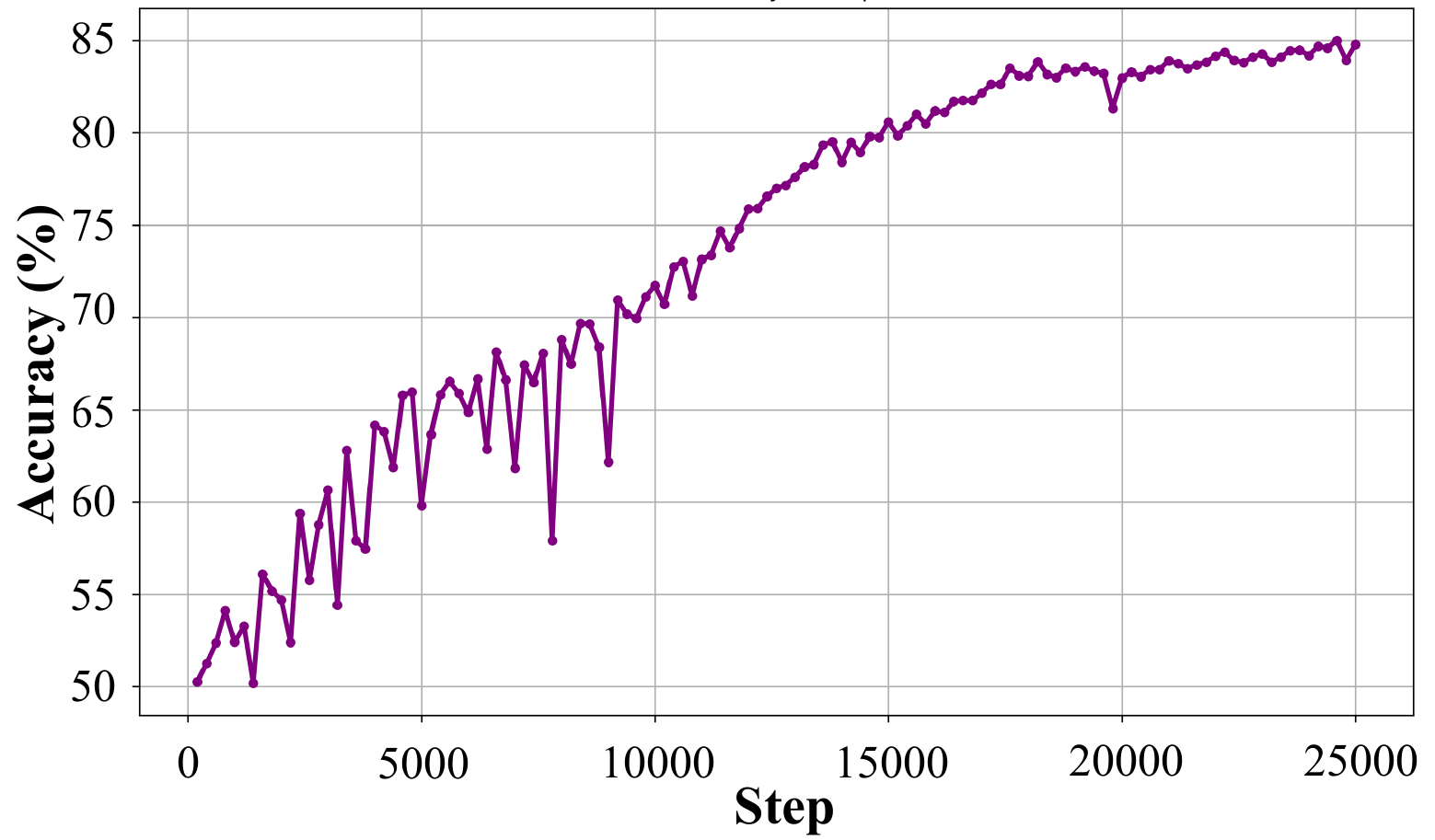}
    \caption{Reward model's accuracy across the training. 
    Checkpoints at specific steps are used for RL experiments.}
    \label{fig:reward-model-training}
    \vspace{-4mm}
\end{wrapfigure}
The prediction head produces a scalar output $s$, and we optimize the model using the MSE between $s$ and the corresponding binary label~\citep{liu2024skyworkrewardbagtricksreward,zhang2024restmcts,deng2024q}.
The model learns to predict the absolute helpfulness, 
facilitating further RL, instead of using contrastive learning to compare the relative helpfulness of paired responses.
The learning rate is $10^{-6}$, and the RM is trained for 25,000 steps. 

The evaluation accuracy dynamics of RMs are shown in Figure~\ref{fig:reward-model-training}. 
The best-performing model achieved an evaluation accuracy of 85\%.
Different RM models with varying accuracies are used in subsequent experiments to simulate the scenarios with different levels of reward noises similarly in practical usages.

\subsection{Learning to reason using reward models of varying accuracy}

\begin{figure}[t]
    \centering
    \includegraphics[width=\linewidth]{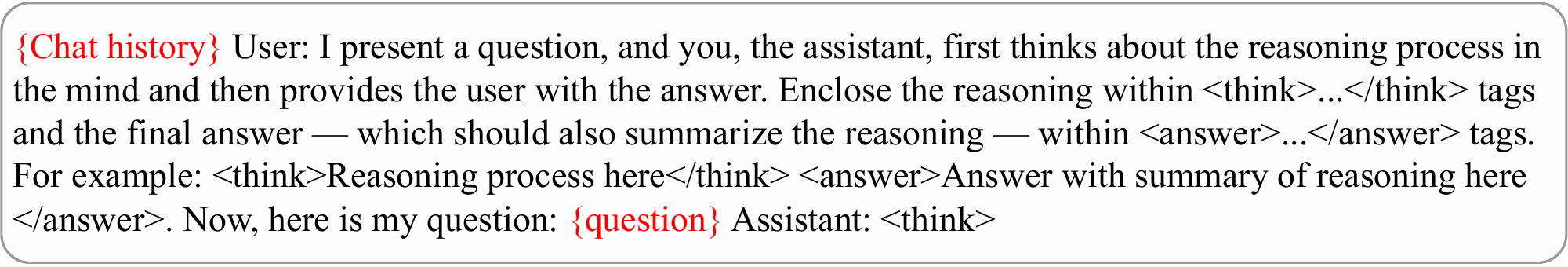}
    \caption{The prompt used in the HelpSteer3 task, where the ``question'' and ``chat history'' placeholders are filled accordingly.
}
    \label{fig:nlp-prompt}
\end{figure}

\textbf{Training.}
The hyperparameters used in this section basically follow those employed in previous math experiments. 
Training is conducted with Qwen-2.5-7B on the HelpSteer3 dataset, lasting a total of 200 steps. 
Figure~\ref{fig:nlp-prompt} shows the prompt template, which instructs the model to first carefully consider how to provide useful assistance. 
It then asks the model to summarize its reasoning and present the final response within the <answer> tag. 
Importantly, the RM only evaluates the text within the <answer> tag, not the entire output.
This approach ensures that the reward pipeline aligns with the one used in the mathematical experiments.

\textbf{Evaluation.} Evaluating open-ended tasks during training presents a much greater challenge than evaluating math problems, due to the lack of objective criteria and the absence of reliable, efficient evaluators. 
Because our most accurate reward model (RM) is used during training, it cannot be employed for evaluation at test time, as LLMs may learn to hack its preferences.
It is also prohibitive to ask more advanced LLMs like ChatGPT or human evaluators to frequently evaluate models during training.
As a result, we perform evaluation only after training, using a subset of 200 samples from the evaluation set, assessed by both GPT-4o and human evaluators.
Specifically, we compare two models by having GPT-4o and human evaluators assess their responses to the same question.
Only text in <answer> tags is used for evaluation.

The prompt used for GPT's evaluation is shown in Figure~\ref{fig:eval-prompt}.
The evaluation considers factors including helpfulness, informativeness, reasoning, and coverage of user needs.
To avoid bias from positional preferences~\citep{liu-etal-2024-lost,chen-etal-2024-fortify,zhang2024middlelanguagemodelsuse} in language models, ChatGPT-4o evaluates an output pair twice for the same question, each time with a different order.
A model's response may result in a win, loss, or tie relative to the other model's response, with the results presented in pie charts. 
In the main text, we report GPT evaluation scores, as they are more reproducible for the community.
Details on human evaluation—guidelines, results, and inter-evaluator agreement measured by Fleiss' Kappa~\citep{fleiss1971mns}—are provided in Appendix~\ref{apx:human}. 
There, we show that human evaluation aligns with GPT assessments, with evaluators demonstrating moderate to substantial agreement.

\begin{wrapfigure}{r}{0.45\linewidth}
    \includegraphics[width=\linewidth]{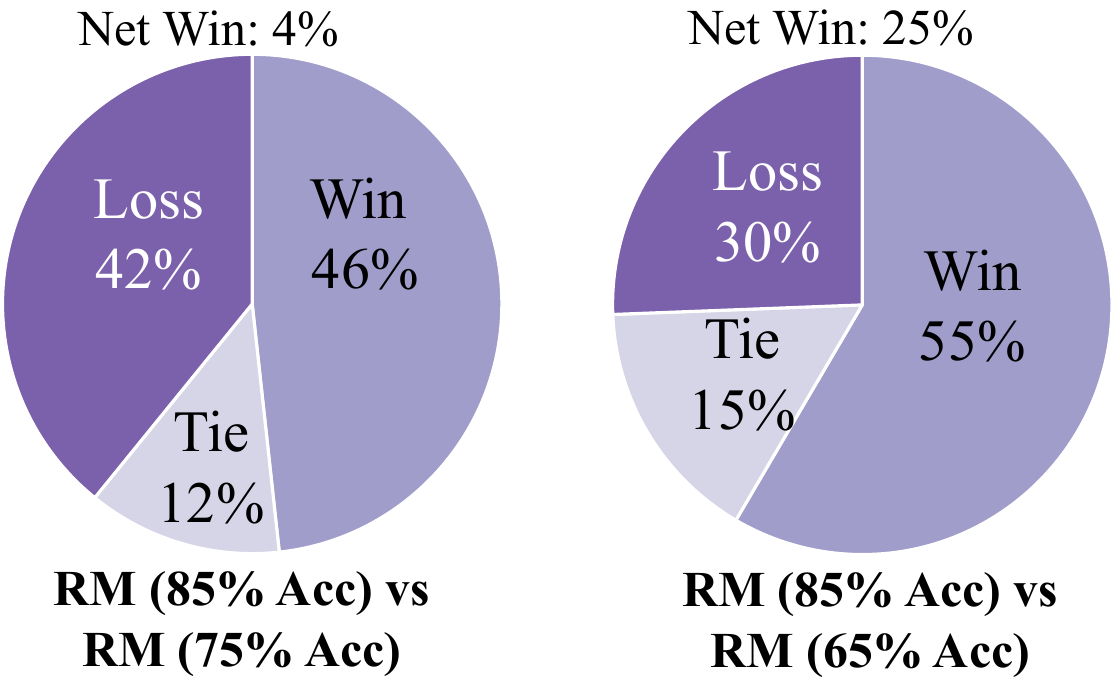}
    \caption{Qwen-2.5-7B trained with an 85\%-accurate RM performs similarly to using a 75\%-accurate RM, but significantly better than using a 65\%-accurate RM.
    The ``Net Win'' refers to the performance advantage of the former RM over the latter.}
    \label{fig:diff-compare}
\end{wrapfigure}
\textbf{\textit{Experiment 3.}}
We compare the performance of the Qwen-2.5-7B model trained with reward models (RMs) of varying accuracies: 85\%, 75\%, and 65\%. 
The results are presented in Figure~\ref{fig:diff-compare}. 
The 85\%-accurate RM yields only a modest 4\% net win-rate advantage over the 75\%-accurate RM, suggesting that their performances are similar. 

Notably, the LLM trained using the 65\%-accurate RM shows a significant decline in downstream performance.
This decline can be attributed to multiple factors:

First, while one might expect the 85\% RM to be just 1.31 times (i.e., 85/65) better than the 65\% RM based on raw accuracy, the actual difference in the number of misclassified labels is more than double (35\% vs. 15\%).
This nonlinear increase in noise significantly impacts the quality of the training signal.

Second, beyond accuracy, the magnitude and distribution of reward scores are also critical.
In contrast to domains like math problem solving—where rewards are typically binary (e.g., 0 or 1)—RMs, especially less accurate ones, tend to produce scores clustered around 0.5, even when they make correct classifications. 
This clustering reflects underlying model uncertainty.
This effect is exhibited in the reduced variance of reward outputs from lower-accuracy RMs: on a validation set, the score variances are 0.1937, 0.1161, and 0.0672 for the 85\%, 75\%, and 65\%-accurate RMs, respectively.
A more accurate RM pushes scores further from the decision boundary, helping to avoid both over- and under-estimated rewards.
These observations align with findings by~\citep{razin2025makesrewardmodelgood}, who emphasized that higher variance is also a key factor in RM effectiveness.
In summary, both the lower accuracy and lower variance of the 65\%-accurate RM likely contribute to its weak downstream performance. 
However, disentangling the individual effects of these factors remains a challenge, as training RMs with both targeted accuracy and targeted variance for clean ablations is difficult.

Nonetheless, our results demonstrate that the Qwen-2.5-7B model performs comparably when trained with reward models that are 75\% and 85\% accurate, indicating a degree of robustness to reward noise—though this robustness is less pronounced than what has been observed in mathematical tasks.

\begin{mytheorem}
\textit{\textbf{Takeaway 3.}} 
While reward noise in neural reward models arises from multiple factors, the robustness to such noise observed in mathematical tasks persists—albeit to a different extent—in open-ended tasks.
\end{mytheorem}

We now understand that effective reasoning does not necessarily require using reward models (RMs) with the highest possible accuracy. 
This insight may offer some relief for real-world applications, where researchers often worry that their RMs are not sufficiently accurate. 
However, as demonstrated by our RM with 65\% accuracy, some noisy RMs are indeed inadequate for practical use as the sole source of reward.
Recognizing the importance of reasoning patterns, we propose a method for calibrating RMs with RPR (Section~\ref{sec:math-exp}). 
This approach overcomes the performance ceiling imposed by the limitations of the reward models at hand.

\subsection{Calibrating noisy RMs with reasoning pattern reward}
\label{sec:nlp-exp}

\textbf{Method.} 
Considering that (1) it is impossible to train a perfect RM, and (2) exploring effective reasoning patterns is crucial for LLMs, we wonder whether reasoning pattern reward (RPR, Section~\ref{sec:math-exp}) could help calibrate noisy RMs and thus obtain better performance.
There are two potential ways to calibrating noisy rewards based on the value of reasoning patterns:

\textbf{1. Compensatory reward for underestimated responses.} When an RM produces false negatives, assigning a low score to an ``objectively'' good response, we give it some compensation.
\footnote{We acknowledge that there are no objective rules in this task; By ``objectively,'' we refer to whether rewarding the response eventually improves performance on the test set.
If it does, the response should be considered good, at least from a deep learning perspective.}
We assume that responses that display better reasoning patterns are likely to be closer to the ``objectively'' good ones.
Therefore, we reuse the RPR as the compensation reward.

\textbf{2. Discounting for overestimated responses.} Conversely, when an RM provides false positive results, that is, it incorrectly assigns a high score to a ``objectively'' poor response, we can apply a discount to RM scores. 
However, this situation is more complex than discounting false negatives. 
The main challenge is determining the appropriate discount factor. 
For instance, if a response receives a full score but lacks key reasoning phrases, should its reward be near zero? 
Setting it too low could overemphasize reasoning pattern rewards, leading to overthinking and performance collapse, as discussed in Section~\ref{sec:math-exp}.
This remains an open research question: how can we effectively calibrate an RM when the noisy reward is a false positive?

\begin{wrapfigure}{r}{0.46\linewidth}
    \vspace{-5mm}\includegraphics[width=\linewidth]{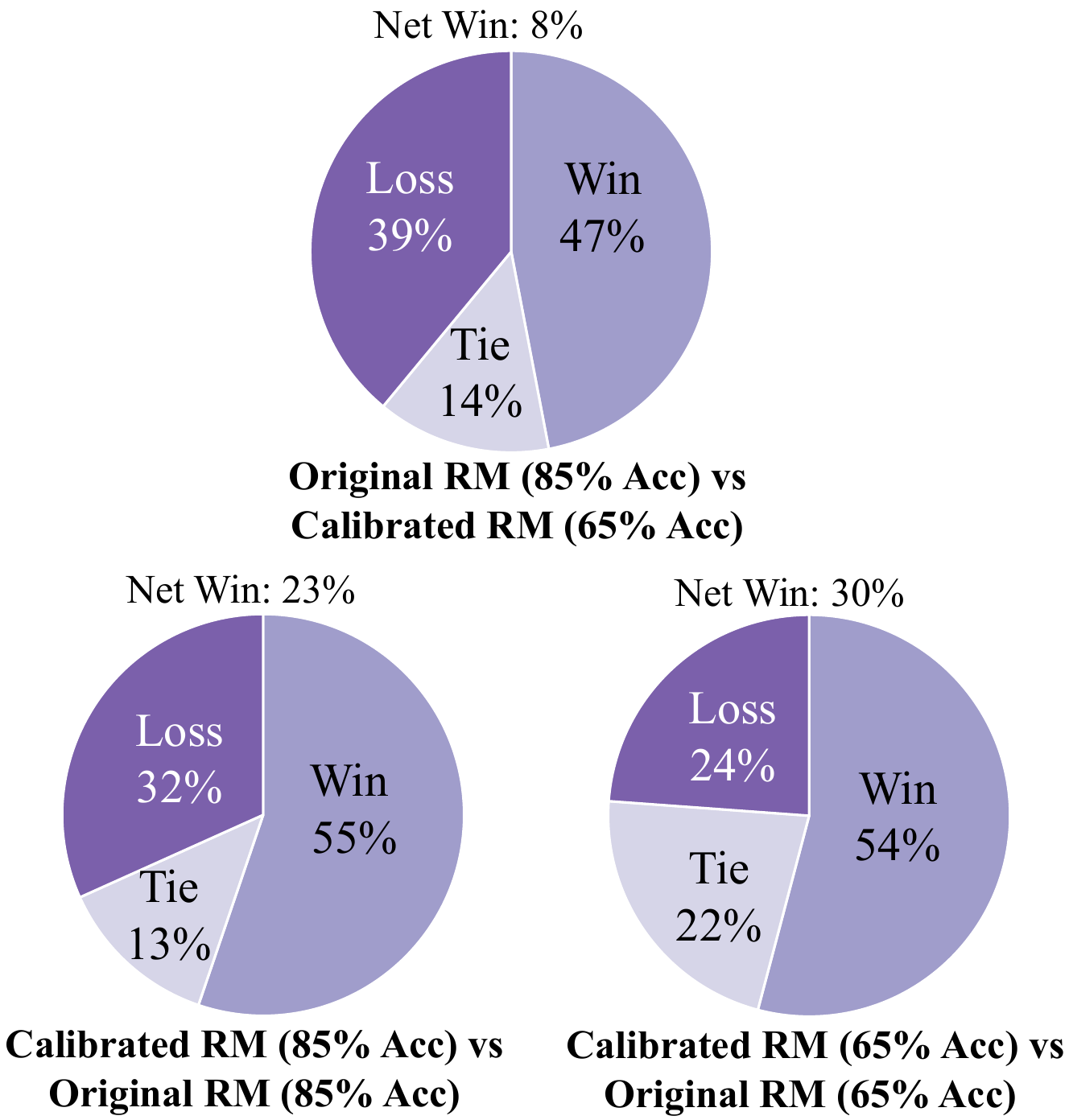}
    \caption{Reward noise calibration effectively enhances downstream performance.}
    \label{fig:self-compare}
    \vspace{-5mm}
\end{wrapfigure}
Given these considerations, we only introduce the first method to calibrate the RM model: When the RM outputs a low score (as determined by a threshold $\tau$), we calculate an RPR score only for the thought text (enclosed in <think> tags), while text in <answer> tags is not considered.
This RPR score is added to the RM output, scaled by a weight $\alpha$.
This calibration incurs no additional time or memory costs.

Note that in our approach, RPR compensates not only for false negatives but also for true negatives. 
This is not problematic, as we demonstrated in Section~\ref{sec:math-exp} that true negative responses still contain valuable reasoning patterns and are therefore worth rewarding.
Another potential concern is that using RPR as the sole reward signal might lead to performance collapse (Figure~\ref{fig:noise-p}).
However, when RPR is used as an auxiliary signal rather than the sole reward, LLMs are trained effectively without such collapse in the following experiments.

\textit{\textbf{Experiment 4.}} We use reward models with accuracies of 65\% and 85\% to conduct several comparisons:
(1) Calibrating the RMs using RPR, applying it to post-training Qwen models, and comparing their performance with models trained solely with the original RMs.
(2) Comparing a Qwen model trained with a 65\%-accurate RM calibrated by RPR to a model trained with an 85\%-accurate RM.
We set the threshold $\tau$ to 0.5 and $\alpha$ to 0.1.
The choice of $\alpha$ is discussed in Appendix~\ref{apx:rollout}.
The results in Figure~\ref{fig:self-compare} demonstrate the effectiveness of RPR calibration:

1. The calibrated 65\%-accurate reward model lags only 8\% behind the 85\%-accurate model—an improvement from an initial 25\% gap, highlighting the substantial gains achieved through calibration.

2. Calibrating noisy RMs boosts downstream LLM performance, outperforming LLMs trained with original RMs. 
Even the 85\%-accurate RM continues to improve after RPR calibration.
RPR calibration addresses the limitations of RMs at hand.

Notably, the improvements observed are not due to an increase in reward score variance (0.1889 and 0.0653 for the 85\% and 65\%-accurate RMs post-calibration), as variance actually decreases slightly.
We provide output examples from models trained with both the original and calibrated RMs in Appendix~\ref{apx:case}.
Furthermore, in Appendix~\ref{apx:other-models}, we show that the Qwen-2.5-3B model, which fails to emerge strong reasoning capabilities under the original RMs, demonstrates such abilities when trained with calibrated RMs. 
This suggests that reasoning pattern rewards not only raise the performance ceiling constrained by reward models in post-trained language models, but also lower the requirements for pre-trained models, enabling less capable models to exhibit reasoning behavior.

\begin{mytheorem}
\textit{\textbf{Takeaway 4.}} RPR can calibrate noisy reward models, particularly when they provide potentially false negative rewards.
Its downstream success highlights the importance of strong reasoning patterns in improving reasoning ability.
\end{mytheorem}

\textbf{\textit{Remark.}} While our RPR is straightforward and could potentially benefit from a more refined design, our primary goal is to demonstrate the importance of reasoning patterns and their effectiveness in real-world scenarios. 
For this purpose, the current simple RPR is sufficient to convey both points through Experiment 4.

\vspace{-3mm}
\section{Related works}
\textbf{Reward model accuracy.} An accurate reward model was considered crucial for successful RL~\citep{frick2024evaluaterewardmodelsrlhf,lambert-etal-2025-rewardbench,liu2025rmbench,zhou2025rmbcomprehensivelybenchmarkingreward}. 
Even in math tasks where rewards are calculated by verification functions, Yeo et al.~\citep{yeo2025demystifyinglongchainofthoughtreasoning} proposed that it is beneficial to refine the reward function with a fine-grained approach for accurately evaluating math answers, considering factors such as output length, correctness, and repetition.
However, Chen et al.~\citep{chen-etal-2024-accuracy} found that more accurate reward models do not necessarily lead to stronger LLMs in downstream tasks. 
Razin et al.~\citep{razin2025makesrewardmodelgood} argues that high reward variance is also important for making the reward model a good teacher. 
Additionally, Wen et al.~\citep{wen2025rethinking} suggested that relying solely on accuracy does not fully capture the impact of reward models on policy optimization. 

Studying the accuracy of reward models from a noise perspective offers some new insights. 
We argue that an accurate reward model is not always necessary in practice, though calibration to noisy rewards improves evaluation.
We provide the first evidence of LLMs' robustness to significant reward noise.

\textbf{The role of RL in post-training LLMs.}
This paper aligns with recent studies suggesting that pre-trained models already possess the fundamental reasoning abilities needed for complex tasks. 
Yeo et al.~\citep{yeo2025demystifyinglongchainofthoughtreasoning} found that pre-training data often includes long chain-of-thought patterns, establishing a foundation for reasoning. 
Similarly, Yue et al.~\citep{yue2025doesreinforcementlearningreally} noted that base models can perform similarly to RL-post-trained models after multiple attempts at difficult tasks. 
Gandhi et al.~\citep{gandhi2025cognitivebehaviorsenableselfimproving} showed that Qwen models outperform Llama~\citep{llama3} models in downstream tasks post-RL, with Qwen models exhibiting natural reasoning behavior. 
Prior works~\citep{ai2025rethinkingreflectionpretraining} demonstrated that reasoning can emerge during pre-training, with models using a reasoning trigger token like ``wait'' to activate chain-of-thoughts and arrive at the correct answer.

We provide strong evidence that models can achieve peak performance, comparable to those trained with strict verification, by rewarding key reasoning patterns instead of requiring correctness verification. 
While RL post-training has seen significant progress, our findings highlight the continued importance of pre-training in building advanced LLMs. 
From a post-training perspective, this also explains why a small amount of high-quality data~\citep{s1} can enhance reasoning abilities, as the foundational capabilities are already present and need effective triggers.

\vspace{-4mm}
\section{Conclusions}

We studied the reward noise, a practical consideration for real-world post-training of LLMs. 
Our findings show that LLMs are highly robust to significant reward noise. 
Surprisingly, when trained solely with reasoning pattern rewards (RPR), the model achieved peak downstream performance on par with models trained with strict correctness verification and accurate rewards. 
Recognizing the importance of reasoning processes over final answers, we use RPR to calibrate noisy reward models.
RPR reduces false negative rewards and improves LLM performance on open-ended tasks.
In the future, enhancing foundational ability during pre-training continues to be promising, and our findings also provide insights for improving post-training techniques.

\bibliography{main}

\begin{thebibliography}{10}

\bibitem{ai2025rethinkingreflectionpretraining}
Essential AI, :, Darsh~J Shah, Peter Rushton, Somanshu Singla, Mohit Parmar, Kurt Smith, Yash Vanjani, Ashish Vaswani, Adarsh Chaluvaraju, Andrew Hojel, Andrew Ma, Anil Thomas, Anthony Polloreno, Ashish Tanwer, Burhan~Drak Sibai, Divya~S Mansingka, Divya Shivaprasad, Ishaan Shah, Karl Stratos, Khoi Nguyen, Michael Callahan, Michael Pust, Mrinal Iyer, Philip Monk, Platon Mazarakis, Ritvik Kapila, Saurabh Srivastava, and Tim Romanski.
\newblock Rethinking reflection in pre-training, 2025.

\bibitem{aime}
Mislav Balunović, Jasper Dekoninck, Ivo Petrov, Nikola Jovanović, and Martin Vechev.
\newblock Matharena: Evaluating llms on uncontaminated math competitions, February 2025.

\bibitem{overthink}
Xingyu Chen, Jiahao Xu, Tian Liang, Zhiwei He, Jianhui Pang, Dian Yu, Linfeng Song, Qiuzhi Liu, Mengfei Zhou, Zhuosheng Zhang, Rui Wang, Zhaopeng Tu, Haitao Mi, and Dong Yu.
\newblock Do not think that much for 2+3=? on the overthinking of o1-like llms, 2025.

\bibitem{chen-etal-2024-accuracy}
Yanjun Chen, Dawei Zhu, Yirong Sun, Xinghao Chen, Wei Zhang, and Xiaoyu Shen.
\newblock The accuracy paradox in {RLHF}: When better reward models don`t yield better language models.
\newblock In Yaser Al-Onaizan, Mohit Bansal, and Yun-Nung Chen, editors, {\em Proceedings of the 2024 Conference on Empirical Methods in Natural Language Processing}, pages 2980--2989, Miami, Florida, USA, November 2024. Association for Computational Linguistics.

\bibitem{chen-etal-2024-fortify}
Yuhan Chen, Ang Lv, Ting-En Lin, Changyu Chen, Yuchuan Wu, Fei Huang, Yongbin Li, and Rui Yan.
\newblock Fortify the shortest stave in attention: Enhancing context awareness of large language models for effective tool use.
\newblock In Lun-Wei Ku, Andre Martins, and Vivek Srikumar, editors, {\em Proceedings of the 62nd Annual Meeting of the Association for Computational Linguistics (Volume 1: Long Papers)}, pages 11160--11174, Bangkok, Thailand, August 2024. Association for Computational Linguistics.

\bibitem{deepseekai2025deepseekr1incentivizingreasoningcapability}
DeepSeek-AI.
\newblock Deepseek-r1: Incentivizing reasoning capability in llms via reinforcement learning, 2025.

\bibitem{deng2024q}
Yanchen Deng, Chaojie Wang, Zhiyi Lyu, Jujie He, Liang Zeng, Shuicheng YAN, and Bo~An.
\newblock Q*: Improving multi-step reasoning for {LLM}s with deliberative planning, 2024.

\bibitem{fleiss1971mns}
J.L. Fleiss et~al.
\newblock {Measuring nominal scale agreement among many raters}.
\newblock {\em Psychological Bulletin}, 76(5):378--382, 1971.

\bibitem{frick2024evaluaterewardmodelsrlhf}
Evan Frick, Tianle Li, Connor Chen, Wei-Lin Chiang, Anastasios~N. Angelopoulos, Jiantao Jiao, Banghua Zhu, Joseph~E. Gonzalez, and Ion Stoica.
\newblock How to evaluate reward models for rlhf, 2024.

\bibitem{gandhi2025cognitivebehaviorsenableselfimproving}
Kanishk Gandhi, Ayush Chakravarthy, Anikait Singh, Nathan Lile, and Noah~D. Goodman.
\newblock Cognitive behaviors that enable self-improving reasoners, or, four habits of highly effective stars, 2025.

\bibitem{hao2024traininglargelanguagemodels}
Shibo Hao, Sainbayar Sukhbaatar, DiJia Su, Xian Li, Zhiting Hu, Jason Weston, and Yuandong Tian.
\newblock Training large language models to reason in a continuous latent space, 2024.

\bibitem{math}
Dan Hendrycks, Collin Burns, Saurav Kadavath, Akul Arora, Steven Basart, Eric Tang, Dawn Song, and Jacob Steinhardt.
\newblock Measuring mathematical problem solving with the {MATH} dataset.
\newblock In {\em Thirty-fifth Conference on Neural Information Processing Systems Datasets and Benchmarks Track (Round 2)}, 2021.

\bibitem{orz}
Jingcheng Hu, Yinmin Zhang, Qi~Han, Daxin Jiang, Xiangyu Zhang, and Heung-Yeung Shum.
\newblock Open-reasoner-zero: An open source approach to scaling up reinforcement learning on the base model, 2025.

\bibitem{jaques-etal-2020-human}
Natasha Jaques, Judy~Hanwen Shen, Asma Ghandeharioun, Craig Ferguson, Agata Lapedriza, Noah Jones, Shixiang Gu, and Rosalind Picard.
\newblock Human-centric dialog training via offline reinforcement learning.
\newblock In Bonnie Webber, Trevor Cohn, Yulan He, and Yang Liu, editors, {\em Proceedings of the 2020 Conference on Empirical Methods in Natural Language Processing (EMNLP)}, pages 3985--4003, Online, November 2020. Association for Computational Linguistics.

\bibitem{lambert-etal-2025-rewardbench}
Nathan Lambert, Valentina Pyatkin, Jacob Morrison, Lester James~Validad Miranda, Bill~Yuchen Lin, Khyathi Chandu, Nouha Dziri, Sachin Kumar, Tom Zick, Yejin Choi, Noah~A. Smith, and Hannaneh Hajishirzi.
\newblock {R}eward{B}ench: Evaluating reward models for language modeling.
\newblock In Luis Chiruzzo, Alan Ritter, and Lu~Wang, editors, {\em Findings of the Association for Computational Linguistics: NAACL 2025}, pages 1755--1797, Albuquerque, New Mexico, April 2025. Association for Computational Linguistics.

\bibitem{liu2024skyworkrewardbagtricksreward}
Chris~Yuhao Liu, Liang Zeng, Jiacai Liu, Rui Yan, Jujie He, Chaojie Wang, Shuicheng Yan, Yang Liu, and Yahui Zhou.
\newblock Skywork-reward: Bag of tricks for reward modeling in llms, 2024.

\bibitem{liu-etal-2024-lost}
Nelson~F. Liu, Kevin Lin, John Hewitt, Ashwin Paranjape, Michele Bevilacqua, Fabio Petroni, and Percy Liang.
\newblock Lost in the middle: How language models use long contexts.
\newblock {\em Transactions of the Association for Computational Linguistics}, 12:157--173, 2024.

\bibitem{liu2025rmbench}
Yantao Liu, Zijun Yao, Rui Min, Yixin Cao, Lei Hou, and Juanzi Li.
\newblock {RM}-bench: Benchmarking reward models of language models with subtlety and style.
\newblock In {\em The Thirteenth International Conference on Learning Representations}, 2025.

\bibitem{llama3}
AI~@~Meta Llama~Team.
\newblock The llama 3 herd of models, 2024.

\bibitem{s1}
Niklas Muennighoff, Zitong Yang, Weijia Shi, Xiang~Lisa Li, Li~Fei-Fei, Hannaneh Hajishirzi, Luke Zettlemoyer, Percy Liang, Emmanuel Candès, and Tatsunori Hashimoto.
\newblock s1: Simple test-time scaling, 2025.

\bibitem{nakano2022webgptbrowserassistedquestionansweringhuman}
Reiichiro Nakano, Jacob Hilton, Suchir Balaji, Jeff Wu, Long Ouyang, Christina Kim, Christopher Hesse, Shantanu Jain, Vineet Kosaraju, William Saunders, Xu~Jiang, Karl Cobbe, Tyna Eloundou, Gretchen Krueger, Kevin Button, Matthew Knight, Benjamin Chess, and John Schulman.
\newblock Webgpt: Browser-assisted question-answering with human feedback, 2022.

\bibitem{ouyang}
Long Ouyang, Jeff Wu, Xu~Jiang, Diogo Almeida, Carroll~L. Wainwright, Pamela Mishkin, Chong Zhang, Sandhini Agarwal, Katarina Slama, Alex Ray, John Schulman, Jacob Hilton, Fraser Kelton, Luke Miller, Maddie Simens, Amanda Askell, Peter Welinder, Paul Christiano, Jan Leike, and Ryan Lowe.
\newblock Training language models to follow instructions with human feedback.
\newblock In {\em Proceedings of the 36th International Conference on Neural Information Processing Systems}, NIPS '22, Red Hook, NY, USA, 2022. Curran Associates Inc.

\bibitem{tinyzero}
Jiayi Pan, Junjie Zhang, Xingyao Wang, Lifan Yuan, Hao Peng, and Alane Suhr.
\newblock Tinyzero.
\newblock https://github.com/Jiayi-Pan/TinyZero, 2025.
\newblock Accessed: 2025-01-24.

\bibitem{razin2025makesrewardmodelgood}
Noam Razin, Zixuan Wang, Hubert Strauss, Stanley Wei, Jason~D. Lee, and Sanjeev Arora.
\newblock What makes a reward model a good teacher? an optimization perspective, 2025.

\bibitem{gpqa}
David Rein, Betty~Li Hou, Asa~Cooper Stickland, Jackson Petty, Richard~Yuanzhe Pang, Julien Dirani, Julian Michael, and Samuel~R. Bowman.
\newblock Gpqa: A graduate-level google-proof q\&a benchmark, 2023.

\bibitem{gae}
John Schulman, Philipp Moritz, Sergey Levine, Michael Jordan, and Pieter Abbeel.
\newblock High-dimensional continuous control using generalized advantage estimation, 2018.

\bibitem{ppo}
John Schulman, Filip Wolski, Prafulla Dhariwal, Alec Radford, and Oleg Klimov.
\newblock Proximal policy optimization algorithms, 2017.

\bibitem{sheng2024hybridflow}
Guangming Sheng, Chi Zhang, Zilingfeng Ye, Xibin Wu, Wang Zhang, Ru~Zhang, Yanghua Peng, Haibin Lin, and Chuan Wu.
\newblock Hybridflow: A flexible and efficient rlhf framework.
\newblock {\em arXiv preprint arXiv: 2409.19256}, 2024.

\bibitem{kimi}
Kimi Team.
\newblock Kimi k1.5: Scaling reinforcement learning with llms, 2025.

\bibitem{helpsteer}
Zhilin Wang, Jiaqi Zeng, Olivier Delalleau, Daniel Egert, Ellie Evans, Hoo-Chang Shin, Felipe Soares, Yi~Dong, and Oleksii Kuchaiev.
\newblock Dedicated feedback and edit models empower inference-time scaling for open-ended general-domain tasks, 2025.

\bibitem{wen2025rethinking}
Xueru Wen, Jie Lou, Yaojie Lu, Hongyu Lin, XingYu, Xinyu Lu, Ben He, Xianpei Han, Debing Zhang, and Le~Sun.
\newblock Rethinking reward model evaluation: Are we barking up the wrong tree?
\newblock In {\em The Thirteenth International Conference on Learning Representations}, 2025.

\bibitem{qwen2.5}
An~Yang, Baosong Yang, Beichen Zhang, Binyuan Hui, Bo~Zheng, Bowen Yu, Chengyuan Li, Dayiheng Liu, Fei Huang, Haoran Wei, Huan Lin, Jian Yang, Jianhong Tu, Jianwei Zhang, Jianxin Yang, Jiaxi Yang, Jingren Zhou, Junyang Lin, Kai Dang, Keming Lu, Keqin Bao, Kexin Yang, Le~Yu, Mei Li, Mingfeng Xue, Pei Zhang, Qin Zhu, Rui Men, Runji Lin, Tianhao Li, Tingyu Xia, Xingzhang Ren, Xuancheng Ren, Yang Fan, Yang Su, Yichang Zhang, Yu~Wan, Yuqiong Liu, Zeyu Cui, Zhenru Zhang, and Zihan Qiu.
\newblock Qwen2.5 technical report.
\newblock {\em arXiv preprint arXiv:2412.15115}, 2024.

\bibitem{yeo2025demystifyinglongchainofthoughtreasoning}
Edward Yeo, Yuxuan Tong, Morry Niu, Graham Neubig, and Xiang Yue.
\newblock Demystifying long chain-of-thought reasoning in llms, 2025.

\bibitem{yue2025doesreinforcementlearningreally}
Yang Yue, Zhiqi Chen, Rui Lu, Andrew Zhao, Zhaokai Wang, Yang Yue, Shiji Song, and Gao Huang.
\newblock Does reinforcement learning really incentivize reasoning capacity in llms beyond the base model?, 2025.

\bibitem{zhang2024restmcts}
Dan Zhang, Sining Zhoubian, Ziniu Hu, Yisong Yue, Yuxiao Dong, and Jie Tang.
\newblock Re{ST}-{MCTS}*: {LLM} self-training via process reward guided tree search.
\newblock In {\em The Thirty-eighth Annual Conference on Neural Information Processing Systems}, 2024.

\bibitem{zhang2024middlelanguagemodelsuse}
Zhenyu Zhang, Runjin Chen, Shiwei Liu, Zhewei Yao, Olatunji Ruwase, Beidi Chen, Xiaoxia Wu, and Zhangyang Wang.
\newblock Found in the middle: How language models use long contexts better via plug-and-play positional encoding, 2024.

\bibitem{zhou2025rmbcomprehensivelybenchmarkingreward}
Enyu Zhou, Guodong Zheng, Binghai Wang, Zhiheng Xi, Shihan Dou, Rong Bao, Wei Shen, Limao Xiong, Jessica Fan, Yurong Mou, Rui Zheng, Tao Gui, Qi~Zhang, and Xuanjing Huang.
\newblock Rmb: Comprehensively benchmarking reward models in llm alignment, 2025.

\bibitem{pmlr-v202-zhu23f}
Banghua Zhu, Michael Jordan, and Jiantao Jiao.
\newblock Principled reinforcement learning with human feedback from pairwise or k-wise comparisons.
\newblock In Andreas Krause, Emma Brunskill, Kyunghyun Cho, Barbara Engelhardt, Sivan Sabato, and Jonathan Scarlett, editors, {\em Proceedings of the 40th International Conference on Machine Learning}, volume 202 of {\em Proceedings of Machine Learning Research}, pages 43037--43067. PMLR, 23--29 Jul 2023.

\end{thebibliography}
\bibliographystyle{plain}

\appendix

\begin{figure}
    \begin{lstlisting}[language={Python}]
from collections import Counter
def calculate_ngram_repetition_penalty(text, n):
    words = text.split()
    ngrams = [tuple(words[i:i+n]) for i in range(len(words) - n + 1)]
    ngram_counts = Counter(ngrams)
    total_ngrams = len(ngrams)
    repeated_ngrams = sum(1 for count in ngram_counts.values() if count > 1)
    repetition_penalty = repeated_ngrams / total_ngrams if total_ngrams > 0 else 0
    return repetition_penalty

def reasoning_pattern_reward(solution):
    reason_pos = solution.find("Assistant: <think>")
    solution_str = solution[think_pos + len("Assistant: <think> "):]
    score = 0
    solution_str = solution.lower()
    score += float("i need to" in solution_str)
    score += float("we need to" in solution_str)
    score += float("wait" in solution_str)
    score += float("alternatively" in solution_str)
    score += float("let me check" in solution_str)
    score += float("let me see" in solution_str)
    score += float("let's focus on" in solution_str)
    score += float("we know that" in solution_str)
    score += float("we can observe " in solution_str)
    score += float("we can see " in solution_str)
    score += float("let me try" in solution_str)
    score += float("let's try" in solution_str)
    score += float("let us try" in solution_str)
    score += float("first," in solution_str) 
    score += float("firstly," in solution_str)
    score += float("next," in solution_str)
    score += float("finally," in solution_str)
    score += float("let us first" in solution_str)
    score += float("let's first" in solution_str) 
    score += float("let me first" in solution_str)
    score += float("try again" in solution_str) 
    score += float("still not" in solution_str)
    score += float("not working" in solution_str) 
    score += float("not correct" in solution_str)
    score += float("does not work" in solution_str)
    score += float("doesn't work" in solution_str)
    score += float("makes sence" in solution_str)
    score += float("since we" in solution_str) 
    score += float("because we" in solution_str)
    score += float("consequently" in solution_str)
    score += float("as a result" in solution_str) 
    score += float("thus" in solution_str)
    score += float("therefore" in solution_str)
    score += float("hence" in solution_str)
    score += float("so that" in solution_str)
    score += float("thereby" in solution_str)
    score += float("if we" in solution_str)
    score += float("given there" in solution_str)
    score += float("for instance" in solution_str)
    score += float("for example" in solution_str)
    score /= 40
    score -= calculate_ngram_repetition_penalty(solution_str, 20)
    score = max(0, score)
    return score
    
\end{lstlisting}
\caption{An example code of reasoning pattern reward without checking answer correctness.}
    \label{fig:code}
\end{figure}

\begin{figure}
    \centering
    \includegraphics[width=\linewidth]{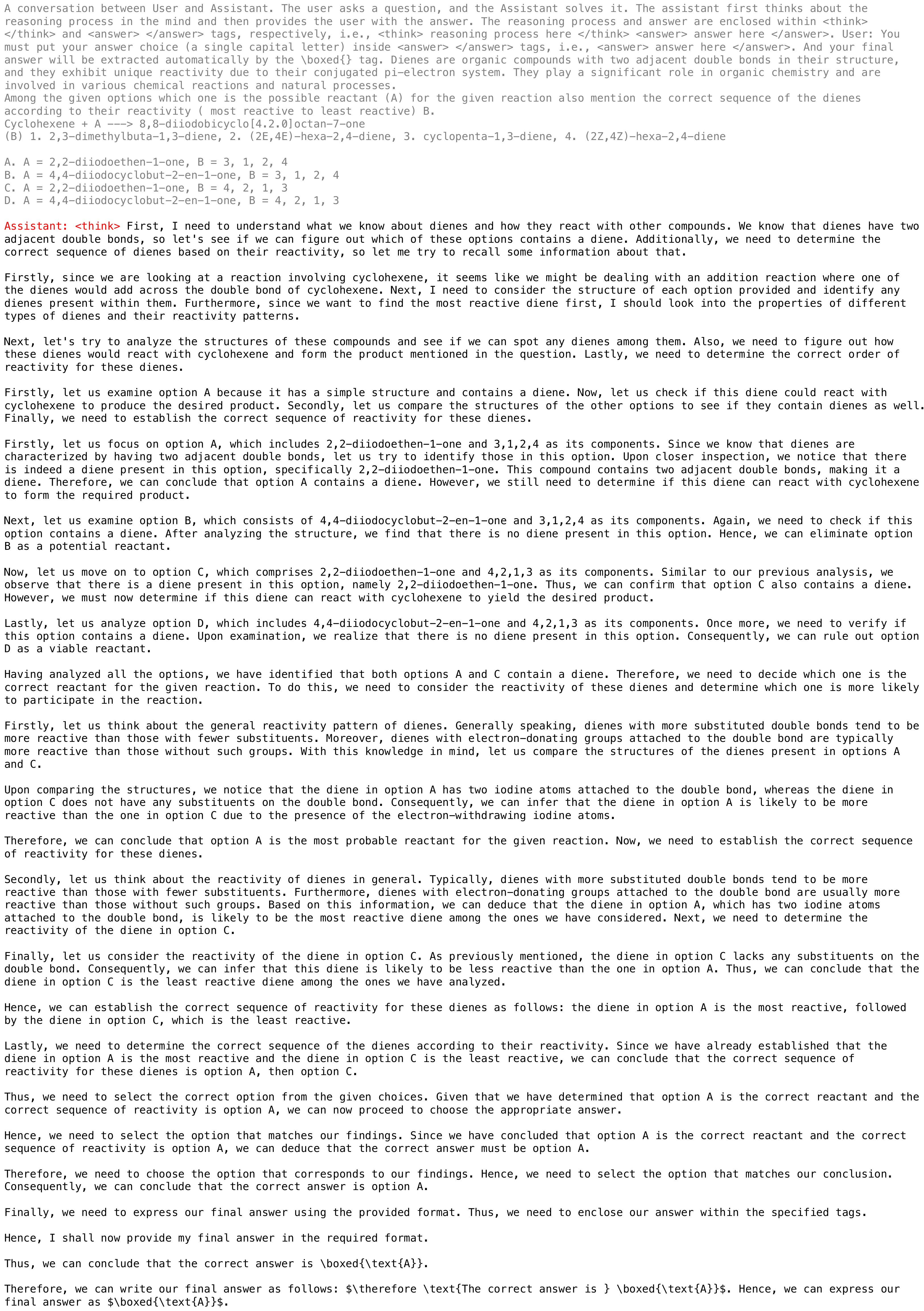}
    \caption{An example of output in the later stage of RL math training, where only the reasoning pattern reward is used without correctness verification. 
    The model has arrived at the correct answer ``A,'' but due to the ongoing reasoning process, the \textcolor{red}{<think>} tag remains open, causing the output length to reach the limit and preventing the correct answer from being generated in answer tags.}
    \label{fig:overthink-case}
\end{figure}

\clearpage
\section{Experiments on Qwen-2.5-3B and Llama-3.1-8B}
\label{apx:other-models}
\begin{figure}[h]
\vspace{-2mm}
    \centering
    \includegraphics[width=0.95\linewidth]{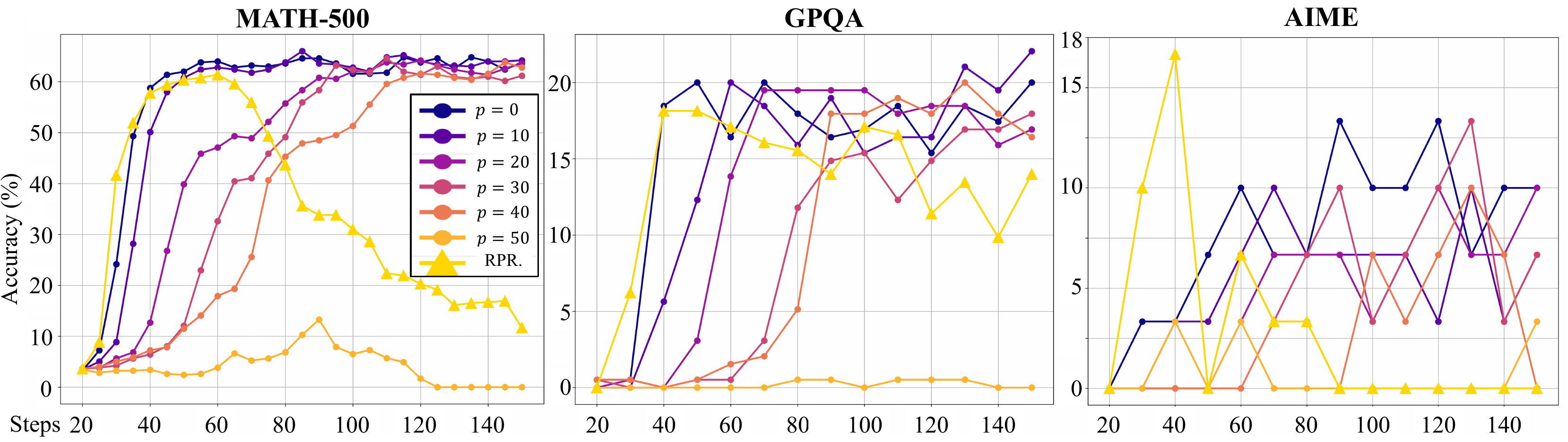}
    \caption{Qwen-2.5-3B results for Experiments 1 and 2: Accuracy on three test sets during training.}
    \label{fig:3b-noise-p}
    \vspace{-2mm}
\end{figure}

In Figure~\ref{fig:3b-noise-p}, we demonstrate that Qwen-2.5-3B exhibits strong robustness to significant reward noise (\textit{\textbf{Experiment 1}}). 
Specifically, it can tolerate up to 40\% of rewards being flipped while still achieving final performance comparable to the noiseless setup. 
However, its convergence under noisy rewards is noticeably slower than that of the 7B models. 
Additionally, performance on the AIME tasks continues to fluctuate, consistent with the behavior observed in the 7B model.

For \textit{\textbf{Experiment 2}}, we find that using RPR as the sole reward signal effectively enables the model to reach peak performance comparable to the noiseless baseline. Notably, on the most challenging AIME tasks, RPR yields the highest peak performance across all setups.

\begin{figure}[h]
    \centering
    \vspace{-4mm}
    \includegraphics[width=0.95\linewidth]{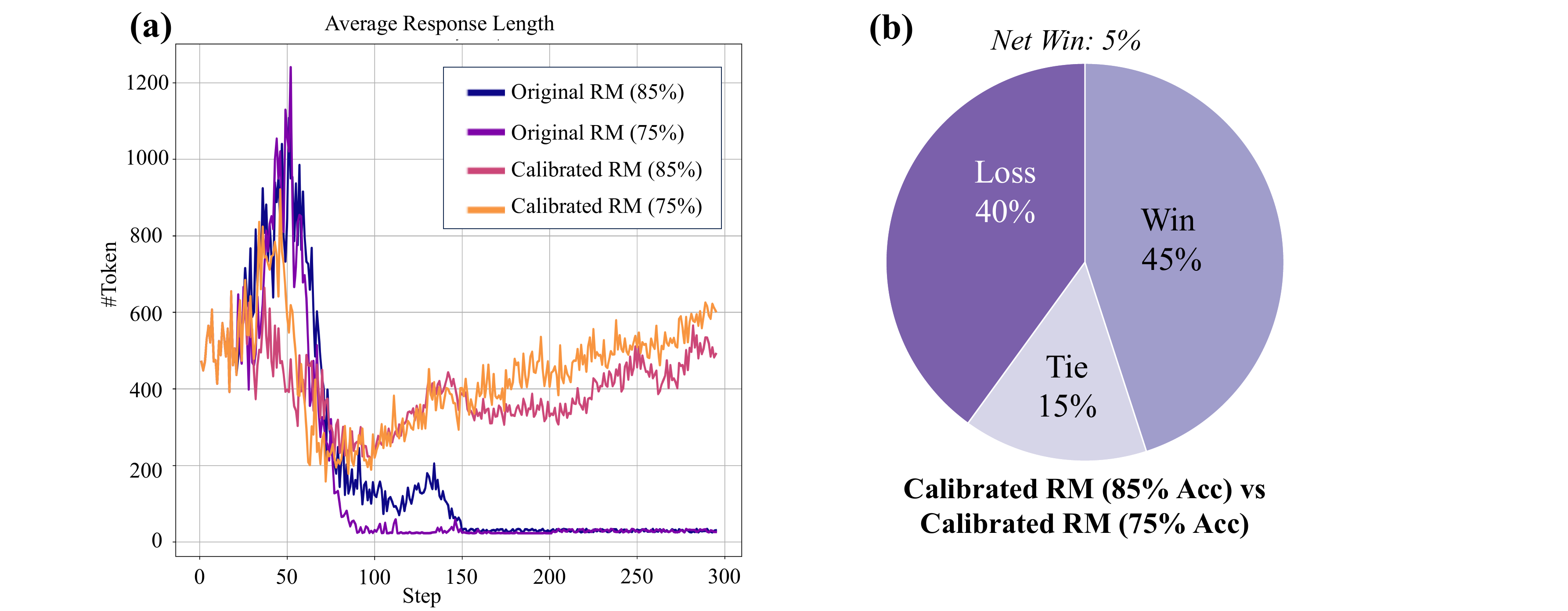}
    \caption{(a) Average response length of Qwen-2.5-3B during training with original vs calibrated RMs. 
    The calibrated RMs successfully enable this small-scale model to perform reasoning, whereas the original RMs fail.
(b) Experiment 3 using Qwen-2.5-3B models trained with calibrated RMs.}
    \label{fig:3b-nlp}
\end{figure}

In the HelpSteer3 task, vanilla RL fails to enable Qwen-2.5-3B to perform effective reasoning. 
We observe that response lengths initially increase but then rapidly collapse to just a few tokens. 
This pattern of ``first-reason-then-collapse'' has also been observed in \citep{tinyzero}, where an LLM is trained to reason on tasks beyond its initial capabilities. 
However, the underlying mechanisms driving this length dynamics remain understudied.
In contrast, when trained with RPR-calibrated RMs (accuracy > 75\%), Qwen-2.5-3B exhibits clear reasoning behaviors.
As shown in Figure~\ref{fig:3b-nlp}(a), the response lengths differ significantly between models trained with original versus calibrated RMs.
These results echo the core insight of \textit{\textbf{Experiment 4}}: calibrated RMs more effectively evoke reasoning abilities in large language models.
Subfigure (b) shows that using an 85\%-accurate RM yields only a 5-point improvement in net win rate over the 75\%-accurate RM—mirroring observations in 7B models in \textit{\textbf{Experiment 3}}.

\begin{figure}[h]
    \centering
    \includegraphics[width=\linewidth]{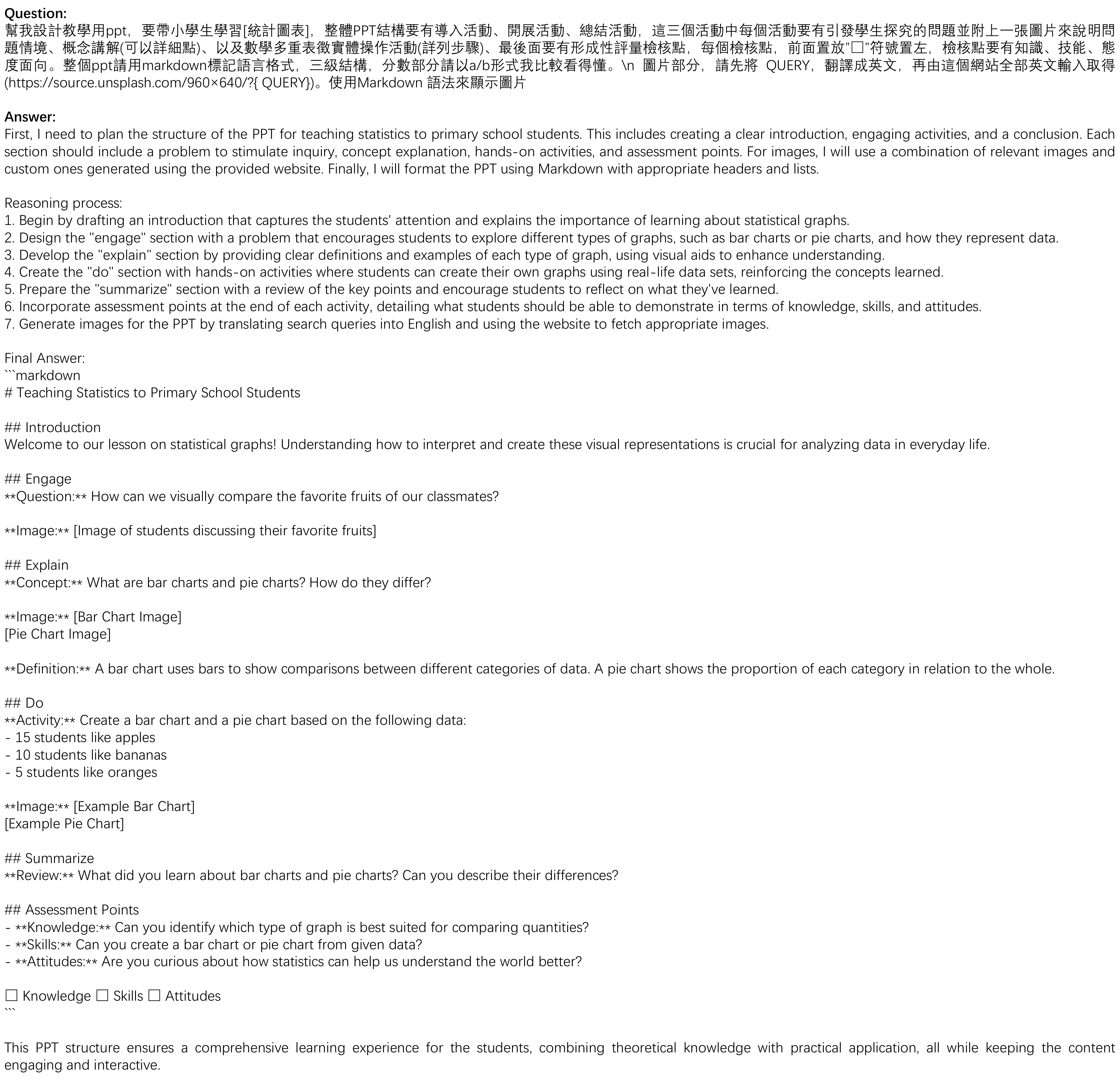}
    \caption{Our calibrated RMs successfully elicit Qwen-2.5-3B's reasoning ability, whereas the original RM fails to do so. 
    This figure presents 1 of 2 output cases.
    The Chinese question translates to: 
``\textcolor{gray}{Please design a teaching PowerPoint for teaching elementary school students about [statistical charts]. The overall structure of the PPT should include three parts: introductory activity, developmental activity, and summary activity.
Each of these activities must include:
A question that triggers student inquiry, along with an image illustrating the context of the problem.
A concept explanation (detailed).
A math hands-on activity using multiple representations, with a clearly listed step-by-step process.
At the end of the PPT, include a formative assessment checklist. Each checklist item should start with the symbol "$\Box$" aligned to the left and should address three aspects: knowledge, skills, and attitudes.
Please present the entire PPT in Markdown format with a three-level heading structure.
For fractions, use the a/b format as it is easier for me to understand.
As for the images, first translate the QUERY into English, and then use the following website with an English query to retrieve the images:https://source.unsplash.com/960x640/?{QUERY}
Use Markdown syntax to display the images.}''}
    \label{fig:3b-nlp-case1}
\end{figure}

\begin{figure}[t]
    \centering
    \includegraphics[width=\linewidth]{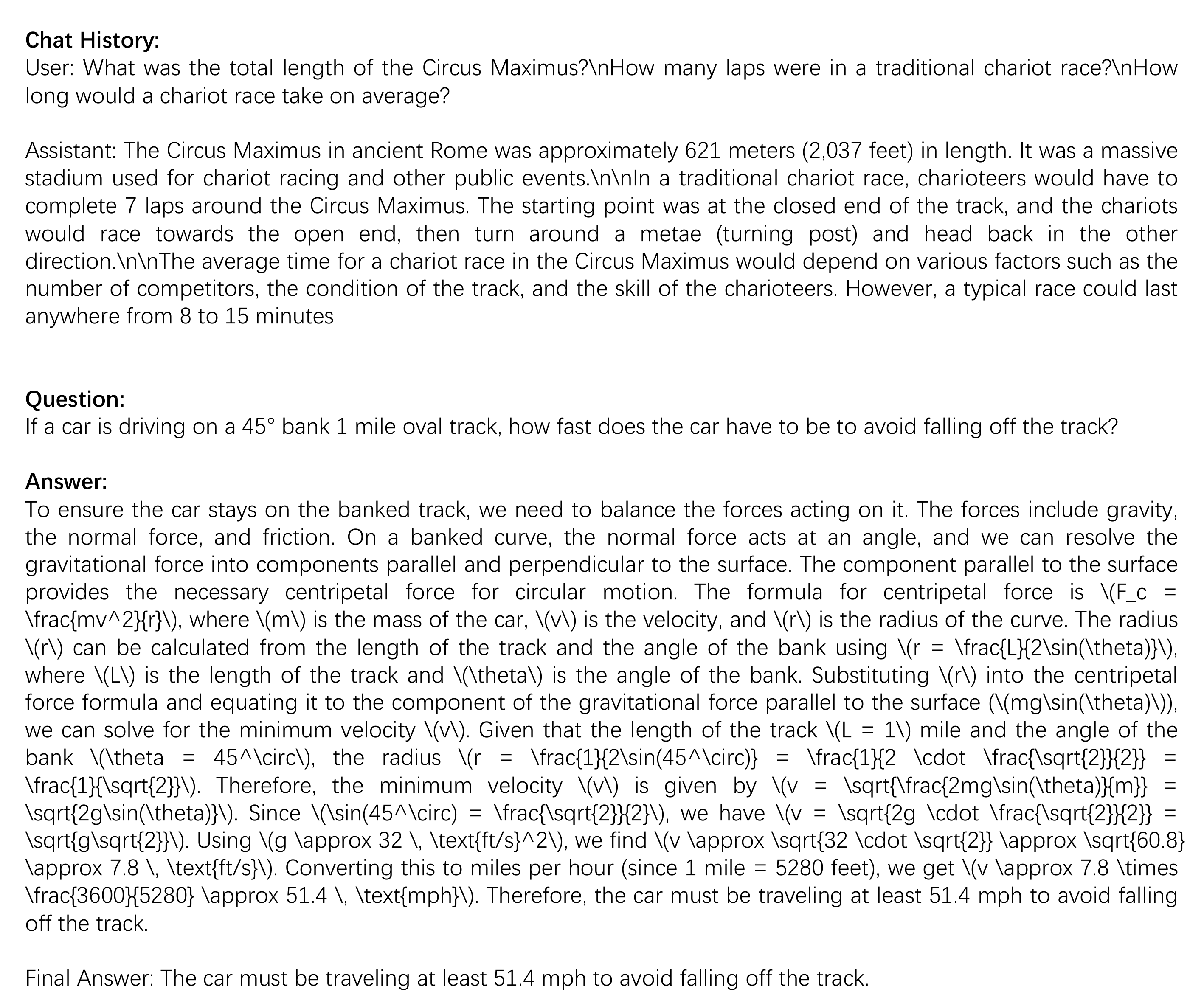}
    \caption{This figure presents the second of two output cases from Qwen-2.5-3B, trained with the calibrated 85\%-accurate RM. 
    The model demonstrates reasoning to solve a physics problem.}
    \label{fig:3b-nlp-case2}
\end{figure}

Figure~\ref{fig:3b-nlp-case1} and Figure~\ref{fig:3b-nlp-case2} present two sample outputs from Qwen-2.5-3B trained with the 85\%-accurate RM, demonstrating that we have successfully elicited the basic reasoning capabilities of this small-scale model, despite some imperfections in these outputs.
In Figure~\ref{fig:3b-nlp-case1}, the Chinese query asks for the creation of a PowerPoint file to teach primary school students about statistical charts.
The 3B model engages in step-by-step reasoning to generate a coherent PowerPoint structure and follows through on its plan.
In Figure~\ref{fig:3b-nlp-case2}, the model processes a complex chat history and solves a physics problem, despite not being explicitly trained for mathematics or physics.

\begin{figure}[h]
    \centering
    \includegraphics[width=\linewidth]{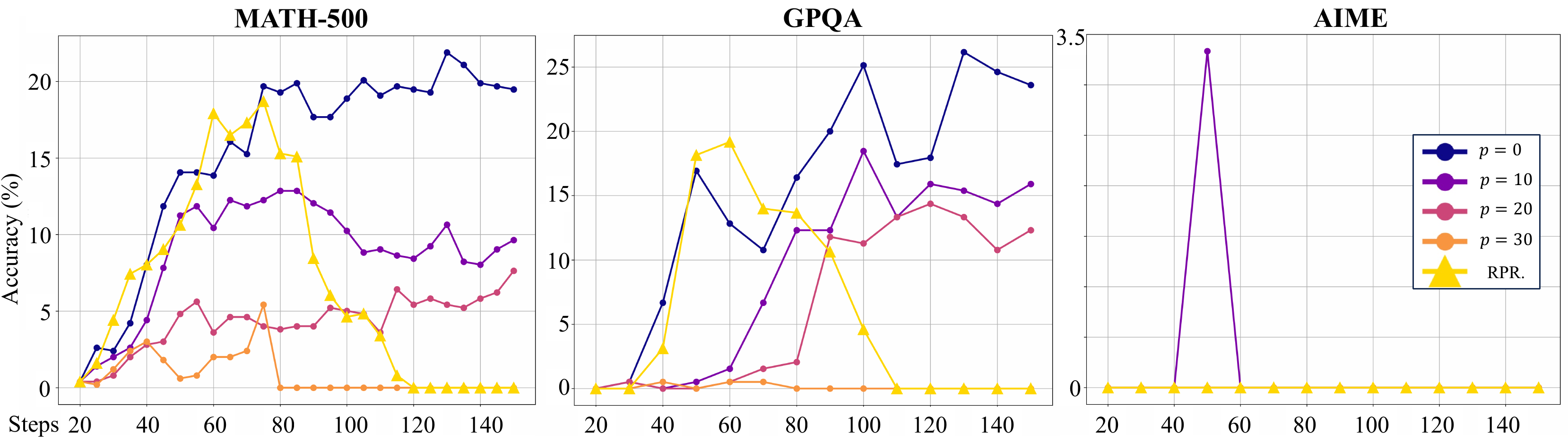}
    \caption{Llama-3.1-8B results for Experiments 1 and 2: Accuracy on three test sets during training (Llama-3.1-8B).}
    \label{fig:llama3-noise-p}
\end{figure}

Llama models are widely recognized to exhibit inherent weaknesses in reasoning capabilities and limited potential for improvement through reinforcement learning~\citep{tinyzero,yeo2025demystifyinglongchainofthoughtreasoning,gandhi2025cognitivebehaviorsenableselfimproving}.
As shown in Figure~\ref{fig:llama3-noise-p}, LLaMA-3.1-8B performs significantly worse than the Qwen models under noiseless conditions and suffers a marked degradation in performance as noise increases.
The accuracy on MATH-500 drops to 0 at $p=30\%$.
Due to its limited foundational capabilities, LLaMA-3.1-8B is not suitable for effective training on the HelpSteer3 task, and thus we omit its results.

\section{Human evaluation}
\label{apx:human}

\begin{figure}[t]
    \centering
    \includegraphics[width=\linewidth]{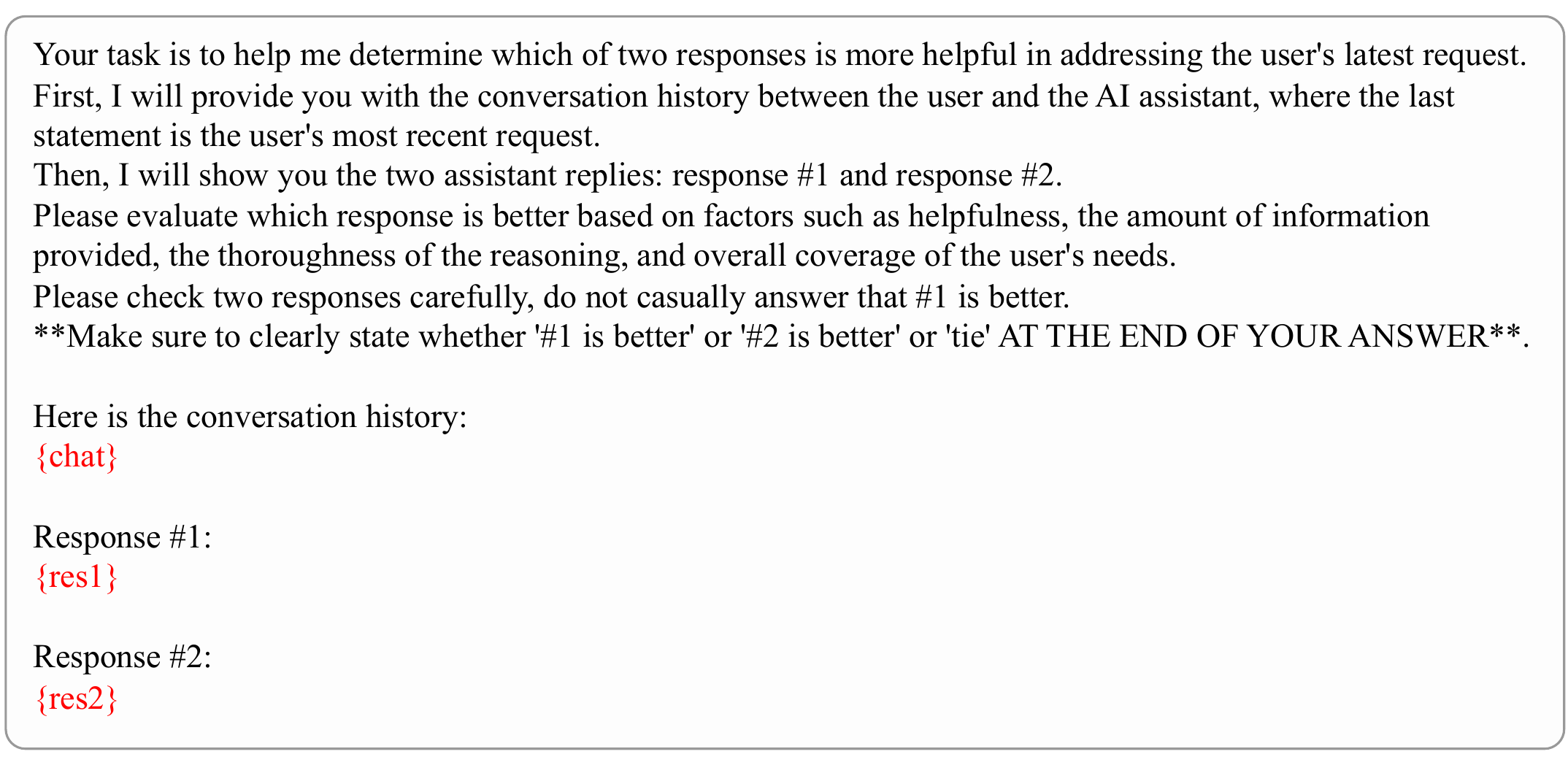}
    \caption{The evaluation prompt for GPT, designed according to the core guidelines for human annotators.
    The placeholders will be replaced with user-assistant chat history and two models' responses.}
    \label{fig:eval-prompt}
\end{figure}

\subsection{Guidelines}
We recruited three graduate students with expertise in model evaluation. 
Each evaluator spent approximately 8 hours completing all tasks and earned \$70 USD. 
The human evaluation was granted by our institute, with the payment slightly above the standard wage for graduate students working in AI companies in our country. 
Below is the guideline for human evaluators:

\textbf{Guideline for Evaluating Responses:}

Your task is to determine which of the two responses better addresses the user's latest request.

\textbf{Steps to Follow:}

\begin{itemize}
    \item \textbf{Review the Conversation History:} Carefully read the conversation history provided. The user's most recent question will be the last message, and that is the request you need to evaluate the responses against.
    \item \textbf{Examine the Two Responses:} You will be presented with two possible replies from two AI assistants (Response \#1 and Response \#2).
    \item \textbf{Criteria for Evaluation:} Evaluate each response based on the following factors:
    \begin{itemize}
        \item \textbf{Helpfulness:} Does the response directly answer the user's request? Is it practical and useful?
        \item \textbf{Amount of Information:}  Does the response provide sufficient details to address the request thoroughly?
        \item \textbf{Clarity and Coherence:} Is the response easy to understand, and does it present information logically?
        \item \textbf{Thoroughness:} Does the response cover all aspects of the user's request? Is anything missing or incomplete?
    \end{itemize}
    \item \textbf{Avoid Quick Judgment:} We will randomize the response order from two models. You cannot infer which one is always better based on the order. Also, don't assume one response is better simply because it's shorter or longer.
\end{itemize}

After evaluating both responses, decide which one is more helpful overall. You can choose \#1 is better, \#2 is better, or they tie with each other. Write the evaluation, as well as reasons.

\subsection{Results and inter-annotator agreement}
\begin{figure}[h]
    \centering
    \includegraphics[width=\linewidth]{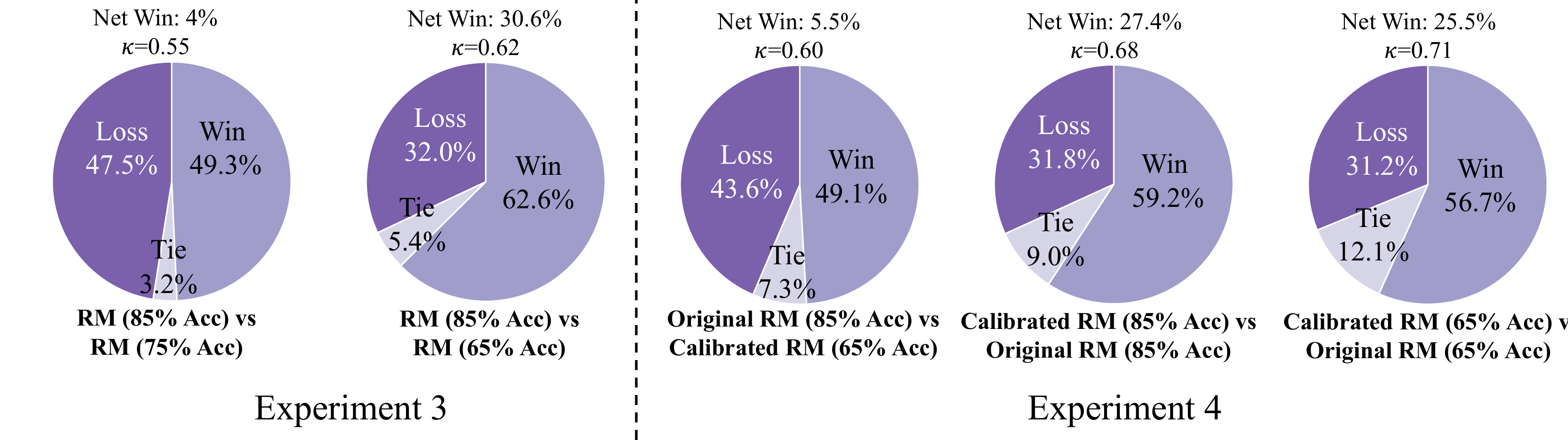}
    \caption{Human evaluation results and agreements.}
    \label{fig:human-compare}
    \vspace{-2mm}
\end{figure}

In Figure~\ref{fig:human-compare}, we present the averaged human evaluation results for Experiments 3 and 4 in Section~\ref{sec:3}. 
Each figure also reports inter-evaluator agreement $\kappa$, with all experiments demonstrating moderate ($0.4<\kappa\leq0.6$) to substantial ($0.6<\kappa\leq0.8$) consistency among evaluators. 
A key distinction between human evaluations and those from GPT-4o is that human judges exhibit stronger discriminatory ability, resulting in fewer comparisons marked as ``ties.'' 
Nonetheless, the overall conclusions—such as the net win ratios and the impact of calibration—align closely with the GPT-based evaluations. 
Therefore, we do not repeat Takeaways and conclusions here.

\section{Case studies}
\label{apx:case}

\begin{figure}
    \centering
    \includegraphics[width=\linewidth]{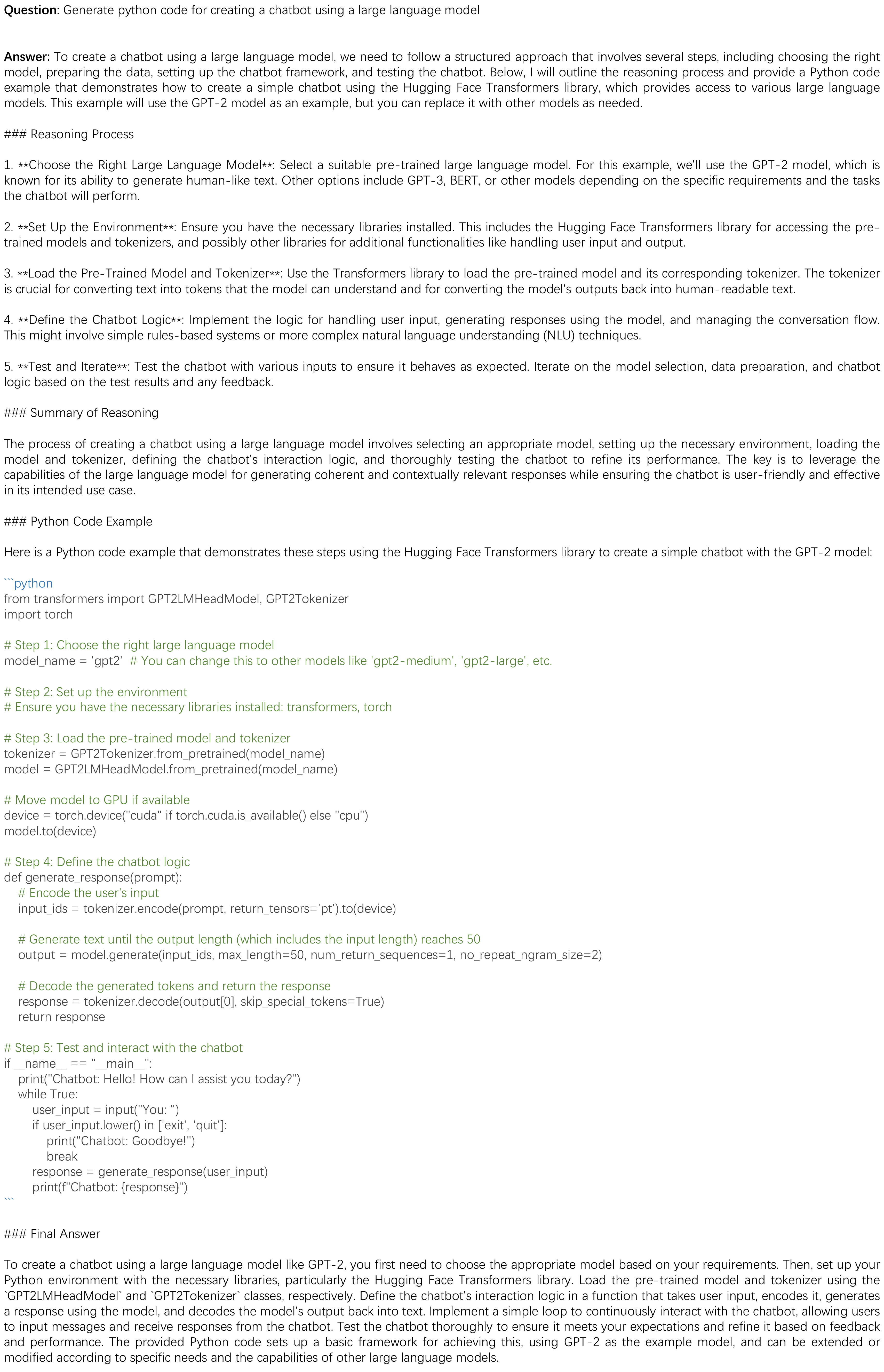}
    \caption{How to create a chatbot using an LLM: the answer from Qwen-2.5-7B trained with the calibrated 85\%-accurate RM.}
    \label{fig:85-think}
\end{figure}

\begin{figure}
    \centering
    \includegraphics[width=\linewidth]{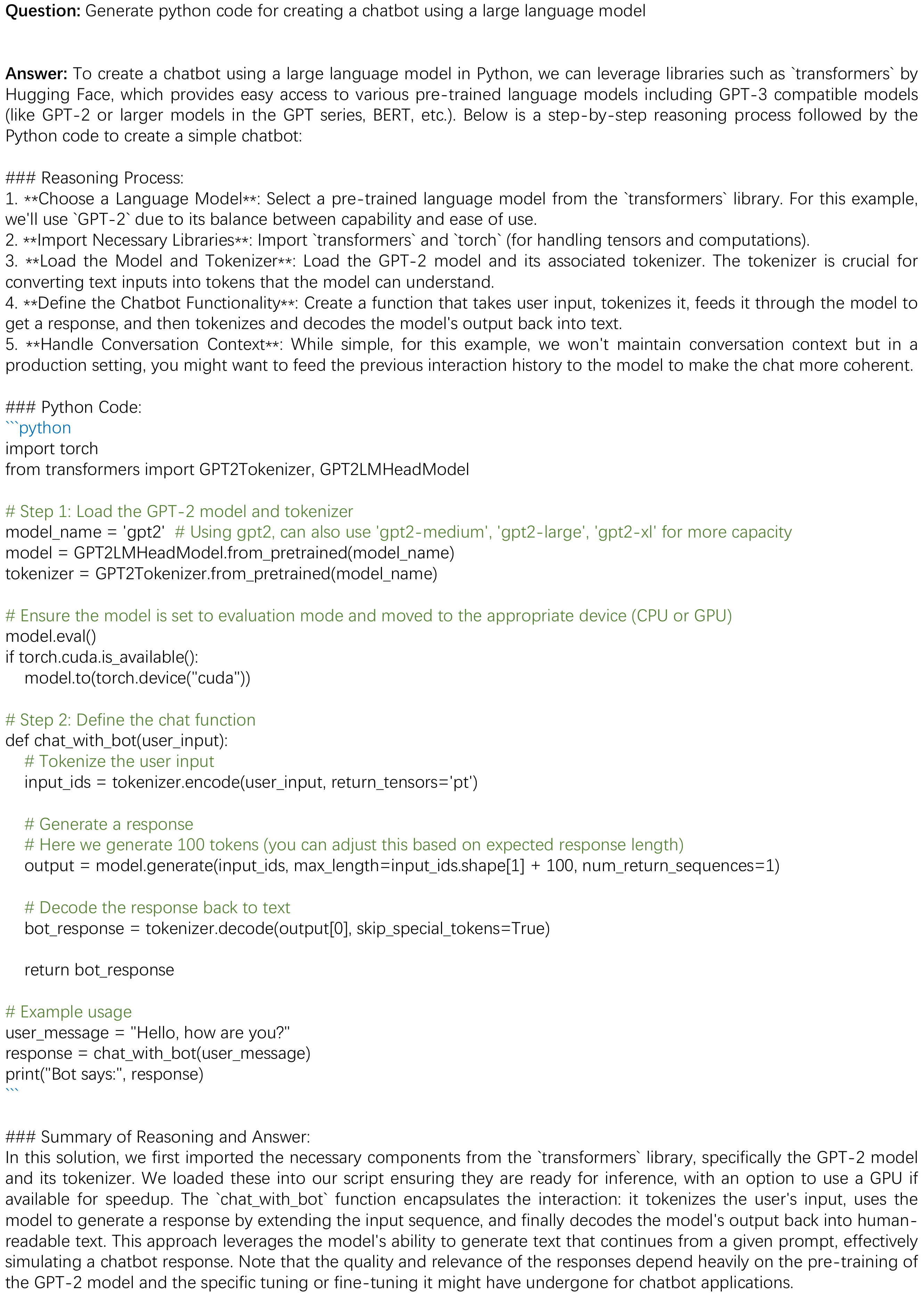}
    \caption{How to create a chatbot using an LLM: the answer from Qwen-2.5-7B trained with the original 85\%-accurate RM.}
    \label{fig:85}
\end{figure}

Figures~\ref{fig:85-think} and \ref{fig:85} show outputs from Qwen-2.5-7B trained with the calibrated and original 85\%-accurate RMs, respectively. 
With RPR, the generated code and comments are more detailed, and both the main function and interaction loop are more comprehensive compared to the single test case produced by models trained without RPR. 
The reasoning process is also more thoroughly articulated.

\begin{figure}
    \centering
    \includegraphics[width=\linewidth]{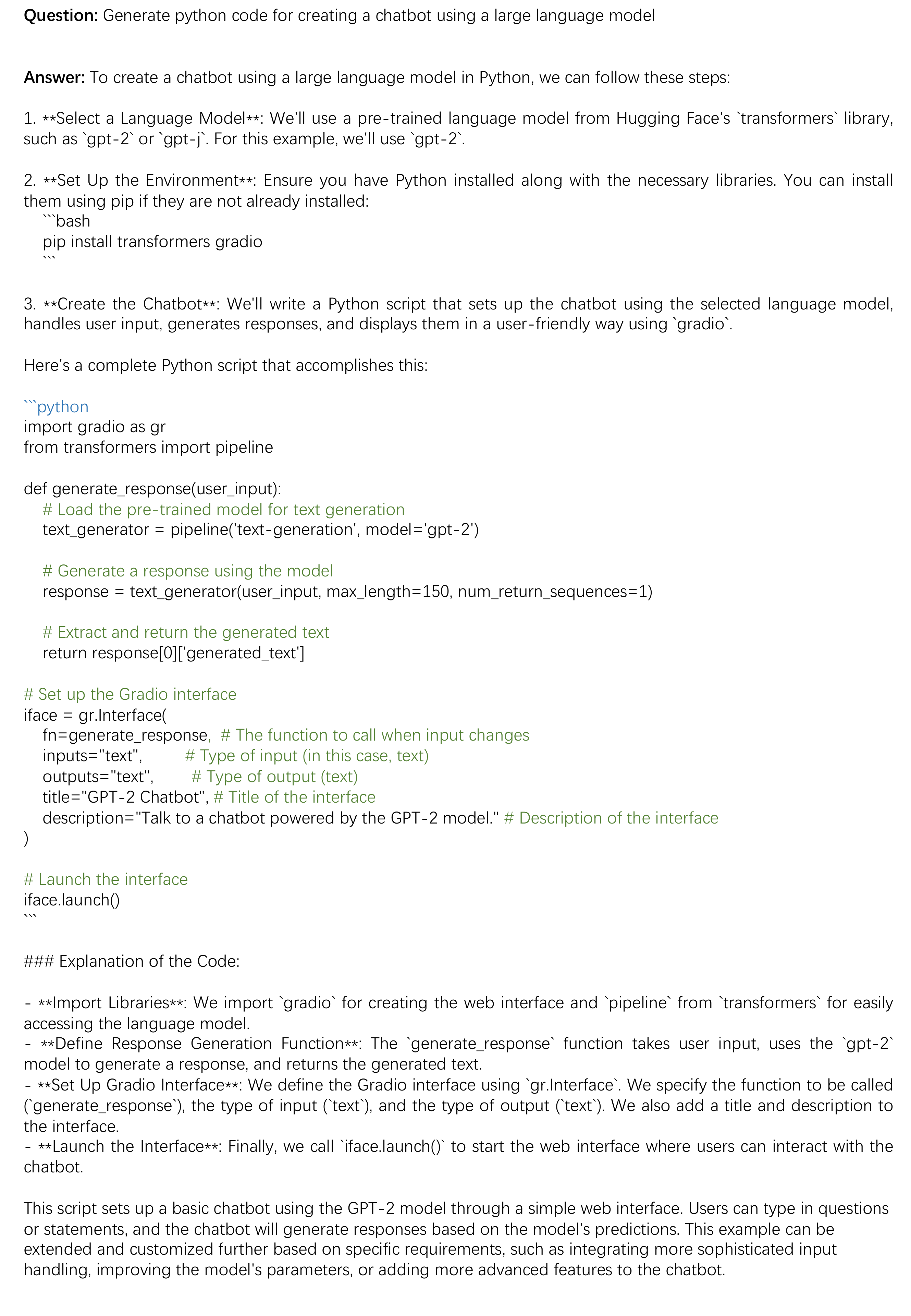}
    \caption{How to create a chatbot using an LLM: the answer from Qwen-2.5-7B trained with the calibrated 65\%-accurate RM.}
    \label{fig:65-think}
\end{figure}

\begin{figure}
    \centering
    \includegraphics[width=\linewidth]{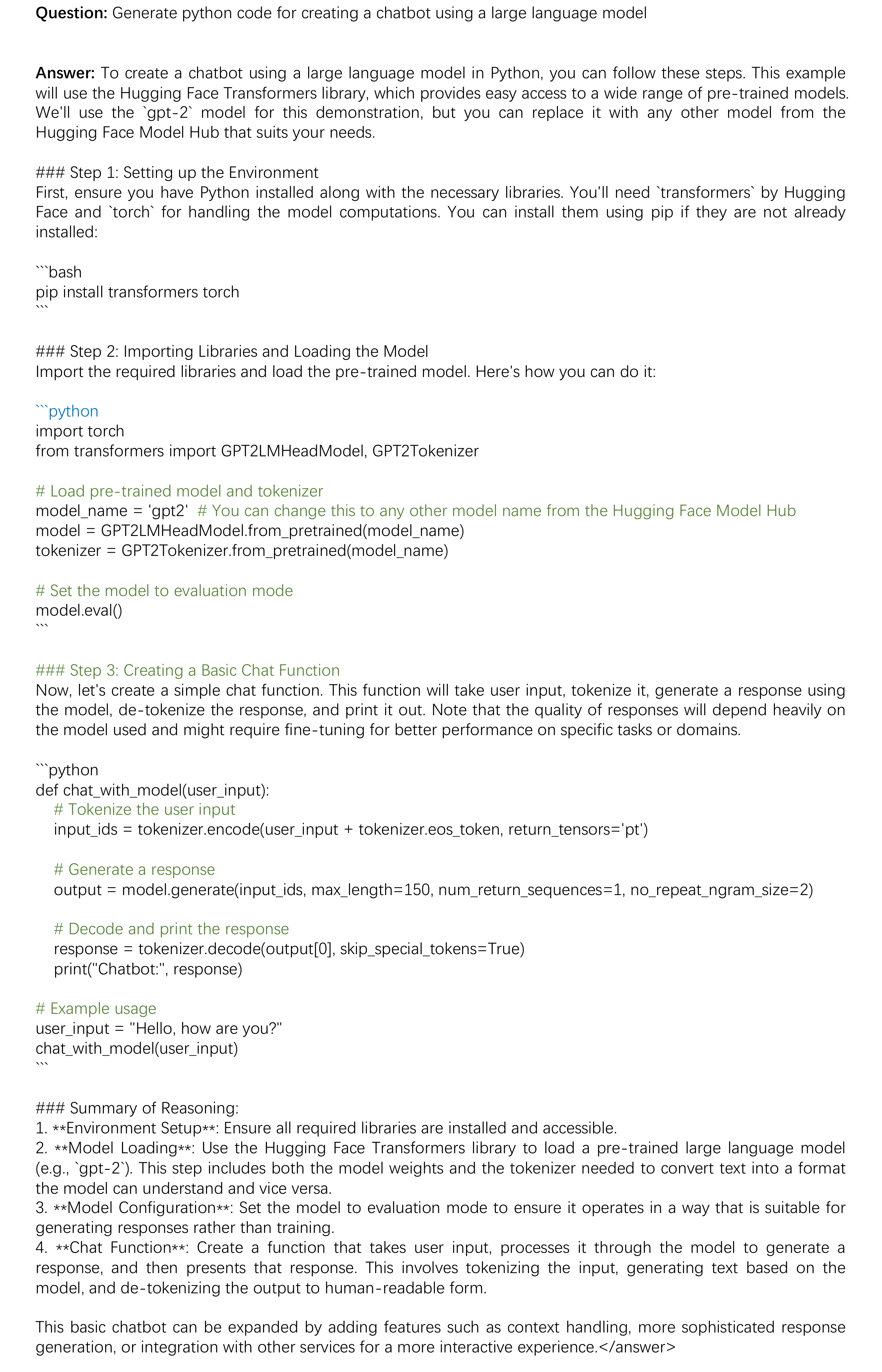}
    \caption{How to create a chatbot using an LLM: the answer from Qwen-2.5-7B trained with the original 65\%-accurate RM.}
    \label{fig:65}
\end{figure}

Figures~\ref{fig:65-think} and \ref{fig:65} illustrate Qwen-2.5-7B trained with calibrated and original 65\%-accurate RMs, respectively. 
Compared to models trained with 85\%-accurate RMs, both outputs here fail to explicitly move the model to the GPU. 
However, the model trained with the calibrated 65\%-accurate RM correctly implements a chatbot using the transformers pipeline API, which implicitly moves the model to the GPU. 
As a result, the model trained with the original 65\%-accurate RM performs slightly worse in terms of helpfulness. 
It is uncommon for an assistant to build a chatbot using the transformers pipeline—an approach that is both concise and effective—suggesting that Qwen models have acquired substantial knowledge during pretraining.

\vspace{-2mm}
\section{Limitations, broader impacts and safety issues}
Due to resource limitations, we limit our experiments to models with up to 7–8 billion parameters. 
However, because our results are consistent even for smaller Qwen-2.5-3B models, and given that the 14B and 70B Qwen models exhibit stronger inherent reasoning abilities, the scalability of our takeaways might not be a concern, as the core premise of these findings relies on strong reasoning capabilities in pre-trained models.

Regarding broader impacts, we hope this paper will inspire future advancements in post-training techniques. 
It is possible that efficient tuning methods could yield similar outcomes with RL. 
Additionally, we highlight the importance of continuing efforts to enhance fundamental reasoning abilities during the pre-training stage.
This paper presents findings and insights into post-training LLMs using RL with noisy rewards, which do not raise safety concerns.
\begin{wrapfigure}{r}{0.42\linewidth}
\vspace{-2mm}
    \includegraphics[width=\linewidth]{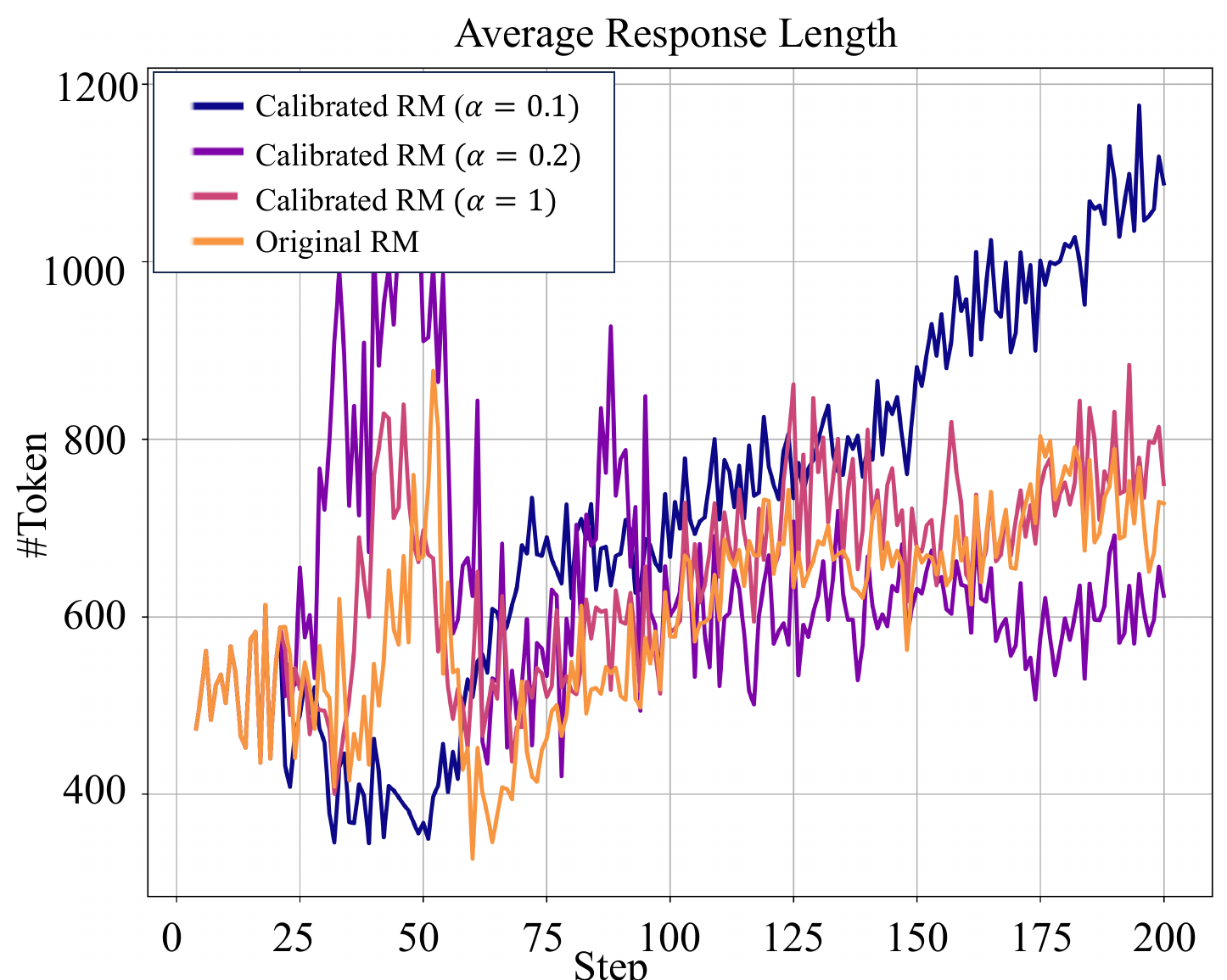}
    \caption{$\alpha=0.1$ enables Qwen's effective reasoning in HelpSteer3 task.
    These experiments use the 85\%-accurate RM.}
    \label{fig:0.1-0.2-1-length}
\end{wrapfigure}

\section{RPR weight in calibration}
\label{apx:rollout}
In Section~\ref{sec:nlp-exp}, we set $\tau = 0.5$ since the RMs are binary classifiers. 
For the scaling factor $\alpha$, we initially used a value of 1, but observed little difference in training dynamics—such as average response length—compared to using the original RMs. 
We then experimented with values of $\alpha$ ranging from 1 to 0.1. 
Notably, reducing $\alpha$ to 0.1 led to a significant increase in average response length (see Figure~\ref{fig:0.1-0.2-1-length}), suggesting a shift in the model's reasoning behavior. 
Based on these findings, we set $\alpha = 0.1$.

\end{document}